\documentclass[letterpaper]{article} 
\usepackage{aaai25}  
\usepackage{times}  
\usepackage{helvet}  
\usepackage{courier}  
\usepackage[hyphens]{url}  
\usepackage{graphicx} 
\usepackage{natbib}  
\usepackage{caption} 
\usepackage{algorithm}
\usepackage{algorithmic}

\usepackage{newfloat}
\usepackage{listings}
\DeclareCaptionStyle{ruled}{labelfont=normalfont,labelsep=colon,strut=off} 
\floatstyle{ruled}
\newfloat{listing}{tb}{lst}{}
\floatname{listing}{Listing}
\usepackage{subcaption}
\usepackage{mathtools}
\usepackage{booktabs}
\usepackage{tabularx}
\usepackage{array}
\usepackage{multicol}  
\usepackage{multirow}
\usepackage{amsmath}
\usepackage{amssymb}
\usepackage{xcolor}
\usepackage{subcaption}

\usepackage{hyperref}

\title{SAFIRE: Segment Any Forged Image Region}
\author {
    Myung-Joon Kwon\textsuperscript{\rm 1\thanks{Equal contribution.}},
    Wonjun Lee\textsuperscript{\rm 1\footnotemark[1]},
    Seung-Hun Nam\textsuperscript{\rm 2},
    Minji Son\textsuperscript{\rm 1},
    Changick Kim\textsuperscript{\rm 1}
}
\affiliations {
    \textsuperscript{\rm 1}School of Electrical Engineering, Korea Advanced Institute of Science and Technology (KAIST)\\
    \textsuperscript{\rm 2}NAVER WEBTOON AI, Seongnam, South Korea\\
    mjkwon2021@gmail.com, dpenguin@kaist.ac.kr, shnam1520@gmail.com, mjson0103@gmail.com, changick@kaist.ac.kr\\
}

\usepackage{bibentry}

\begin{document}

\maketitle

\begin{abstract}
Most techniques approach the problem of image forgery localization as a binary segmentation task, training neural networks to label original areas as 0 and forged areas as 1. In contrast, we tackle this issue from a more fundamental perspective by partitioning images according to their originating sources. To this end, we propose Segment Any Forged Image Region (SAFIRE), which solves forgery localization using point prompting. Each point on an image is used to segment the source region containing itself. This allows us to partition images into multiple source regions, a capability achieved for the first time. Additionally, rather than memorizing certain forgery traces, SAFIRE naturally focuses on uniform characteristics within each source region. This approach leads to more stable and effective learning, achieving superior performance in both the new task and the traditional binary forgery localization. Code: \href{https://github.com/mjkwon2021/SAFIRE}{https://github.com/mjkwon2021/SAFIRE}
\end{abstract}

%
\section{Introduction}
\label{sec:intro}

In the era of artificial intelligence (AI), the proliferation of image editing software~\cite{fu2023guiding,yu2023inpaint} and sophisticated generative models~\cite{rombach2022high,ho2020denoising} has made image forgery more accessible and more challenging to detect than ever before~\cite{lin2024detecting}. 
The ease of image manipulation critically affects areas where the integrity of visual information is crucial, including the spread of fake news in journalism, the use of counterfeit evidence in law enforcement, and the presence of fabricated microscopy images in biomedical research~\cite{verdoliva2020media, sabir2021biofors}.
Therefore, detecting and precisely localizing forgeries within an image is crucial for maintaining trust in digital media.

\begin{figure}[t]
\centering{\includegraphics[width=1.0\linewidth]{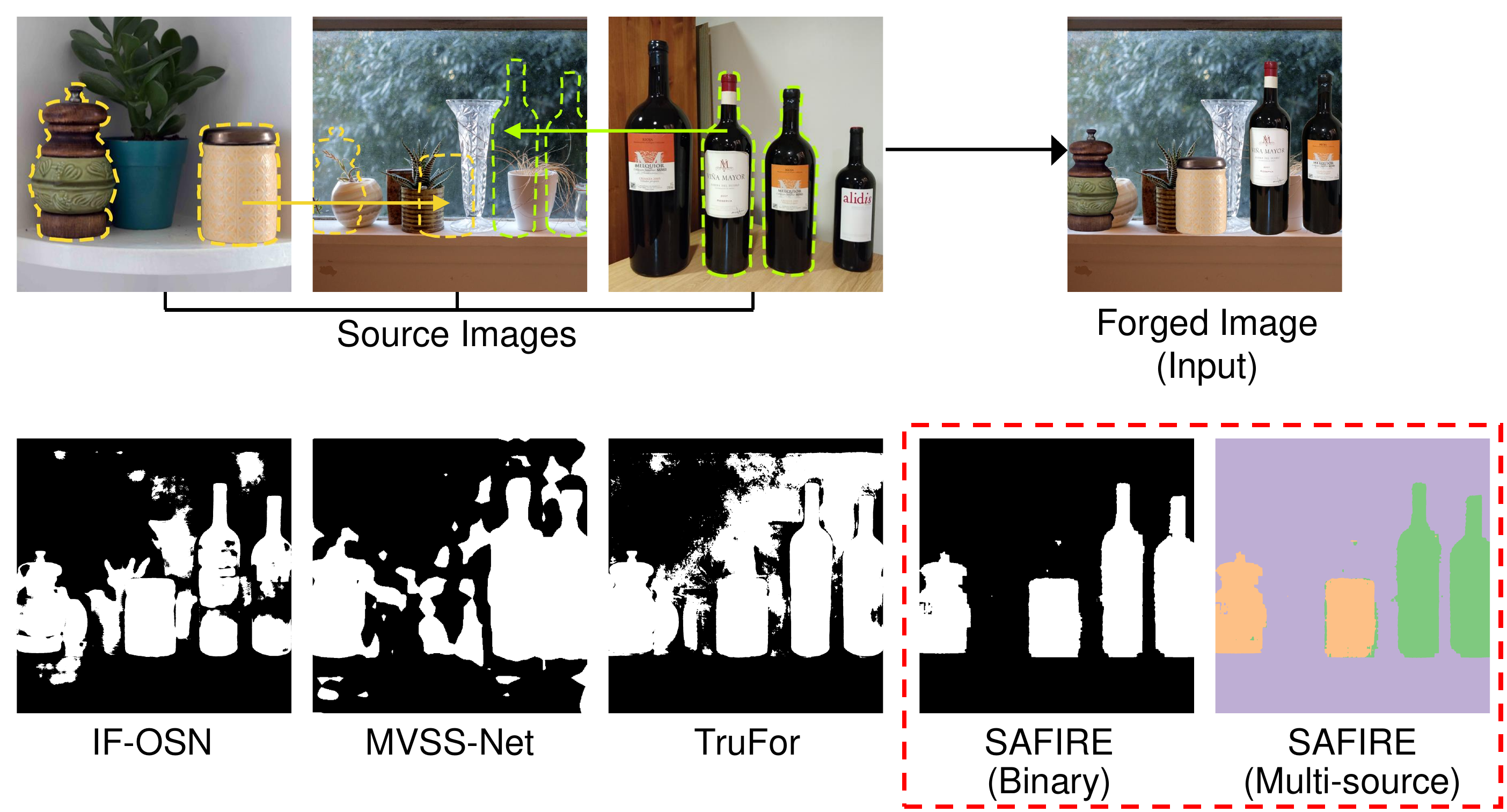}}
\caption{
The forged image is composed of three source regions.
Previous methods are limited to binary prediction --- segmenting forged regions. In contrast, our SAFIRE is also capable of multi-source prediction --- distinguishing regions that originate from the same source images.
}
\label{saf.fig.main_pred}
\end{figure}

\begin{figure}[t]
\centering{\includegraphics[width=1.0\linewidth]{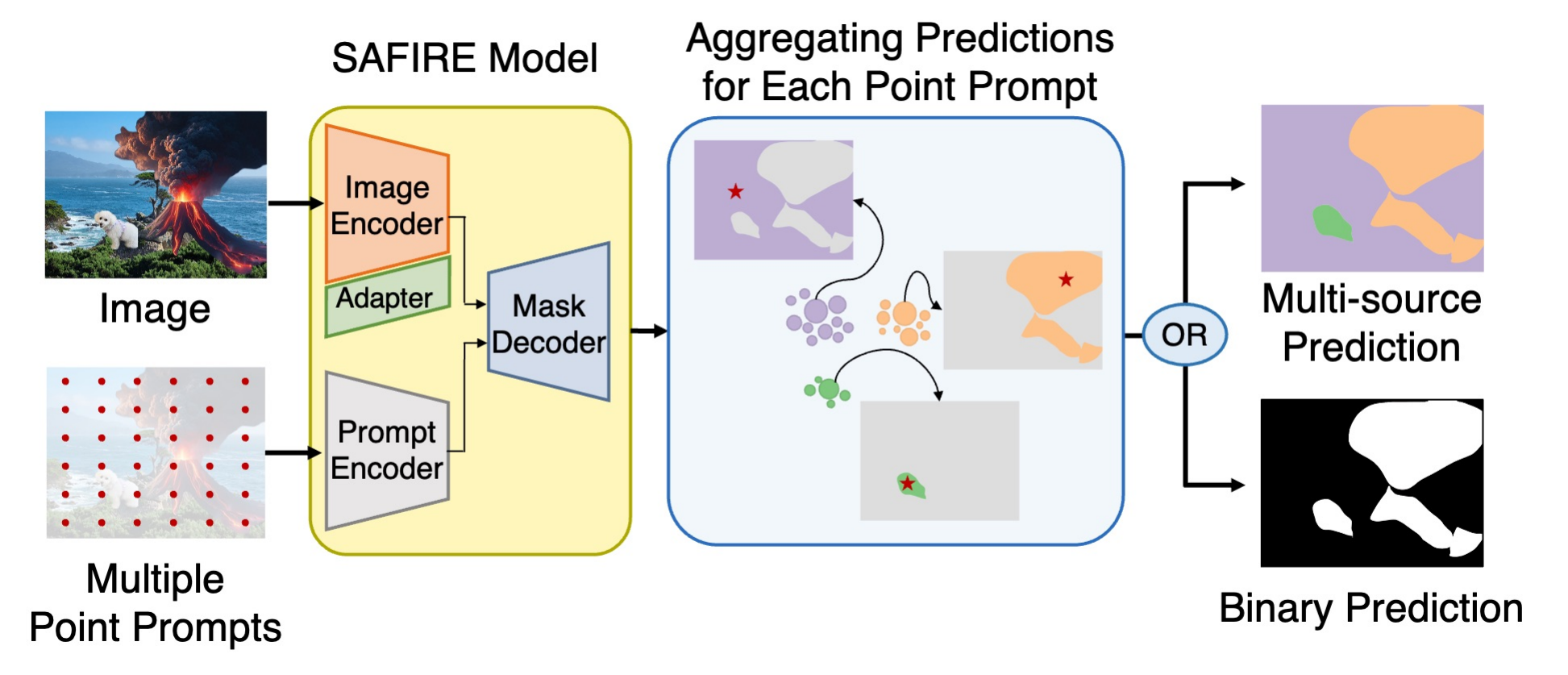}}
\caption{Overview of how SAFIRE conducts IFL. 
An image along with point prompts are input into the model. The model segments the source region containing each point, and these results are combined to produce the final output
}
\label{saf.fig.main}
\end{figure}

Currently, most image forensics methods address the problem of \textbf{image forgery localization (IFL)} through binary segmentation~\cite{guillaro2023trufor,kwon2022learning,liu2022pscc,dong2022mvss,hu2020span,wu2022robust,zhou2023pre,ji2023uncertainty,sun2023safl}. That is, within an image, regions that remain unchanged from the camera capture are labeled as 0, and regions that have been manipulated are labeled as 1, to train deep neural networks.


Instead, we view the IFL from a more fundamental perspective of \textit{partitioning an image into distinct regions based on their origins}.
In this context, we define these distinct regions as \textbf{source regions}, which are distinct segments of an image that have been independently captured, AI-generated, or manipulated (Fig.~\ref{saf.fig.main_pred}).

From this perspective, we propose \textbf{S}egment \textbf{A}ny \textbf{F}orged \textbf{I}mage \textbf{Re}gion (\textbf{SAFIRE}), a novel point prompt-based IFL method designed to precisely partition images into regions based on their original sources.
SAFIRE employs point prompting, where each point on an image segments the area that shares the same source (Fig.~\ref{saf.fig.main}).

To achieve this, we capitalize on the Segment Anything Model (SAM)'s~\cite{kirillov2023segment} point prompting capability with several differences compared to the original SAM. 
First, SAFIRE segments a source region containing the given point whereas SAM segments any meaningful chunk around the point.
Second, while SAM deals with ambiguous ground truths, SAFIRE has a clear ground truth, where all points on one source region share the same answer.
Third, SAFIRE generates and uses point prompts internally, so there is no need for manual input to designate points.

The SAFIRE framework consists of pretraining, training, and inference phases.
In the pretraining phase, source region-based contrastive learning is applied to enhance the feature extraction ability of the image encoder.
In the training phase, the model is trained to segment the source region corresponding to a given point prompt.
Although the model can be trained using forgery datasets with only binary labels, it can perform multi-source prediction.
During inference, a grid of points generates multiple masks, which are then combined to produce the final source partitioning result.




SAFIRE is the first method capable of distinguishing each source when an image has been forged twice or more, resulting in three or more sources. Differentiating each source provides a better explanation of the manipulated image than simply locating forged pixels. Additionally, it facilitates subsequent analysis, such as provenance filtering, which involves retrieving the donor image for each source region in a set of candidate images~\cite{pinto2017provenance,moreira2018image,verdoliva2020media}. Therefore, multi-source partitioning is particularly beneficial for image forensics in real-world scenarios where multiple manipulations are common.

Furthermore, the novel prompting enables SAFIRE to learn effectively by considering the interchangeable nature of authentic and tampered regions, a characteristic we refer to as label agnosticity in IFL.
Tampered areas often lack common traces and are simply different sources within the image compared to authentic areas~\cite{huh_fighting_2018}.
Consequently, attempts to memorize forgery traces lead to confusion and result in unstable learning.
In contrast, SAFIRE uses points as references to learn the uniform characteristics of each source region, rather than memorizing forgery traces. This approach results in stable and effective learning, achieving high performance in both traditional binary IFL and new source partitioning tasks.

As the first paper to solve IFL through multi-source partitioning, we have created a dataset, \textbf{SafireMS}, composed of multi-source images to promote further research in this area. We plan to make it publicly available.

Our primary contributions can be summarized as follows: 
\begin{itemize}
\item 
We introduce a new IFL task that partitions forged images by each originating source.
It helps in understanding the composition of the forged image and makes further analysis easier.

\item 
We propose SAFIRE, a novel IFL method that uses point prompting internally. It is the first technique capable of multi-source partitioning, yet it can be trained using traditional binary datasets.

\item 
Extensive experiments show that SAFIRE demonstrates top performance in both the traditional binary IFL and the new task.

\item
To facilitate the research on the new task, we construct and release a forgery dataset containing images composed of multiple sources.





\end{itemize}


\section{Related Work}
\label{sec:relwork}

\subsection{Image Forgery Localization}

Effective extraction of forensic clues is essential in IFL. This often involves determining which forensic fingerprints to be utilized. These artifacts, often low-level and inconspicuous, include local CFA artifacts~\cite{bammey2020adaptive}, edge information~\cite{dong2022mvss,li2023edge}, JPEG compression artifacts~\cite{kwon2021cat,kwon2022learning}, unique traces left by different camera models~\cite{guillaro2023trufor}, and explicitly enhanced noise~\cite{zhu2024learning}.

Designing network architectures specifically tailored for forensics is another key component in solving IFL. This includes the application of steganalysis filter~\cite{zhou_learning_2018}, various low-level filters and anomaly-enhancing pooling~\cite{wu_mantra-net_2019}, utilizing both top-down and bottom-up paths~\cite{liu2022pscc}, and efficient modeling of internal relationships using Transformers~\cite{hu2020span,hao2021transforensics,wang2022objectformer,zeng2024mgqformer}.

Learning the common characteristics of an image and using consistency as a criterion is also a viable approach.
The pioneering study~\cite{huh_fighting_2018} trains a model using self-supervised learning to determine if two image patches have the same EXIF metadata and uses clustering to check consistency.
Subsequent research has utilized the consistency of camera model fingerprints~\cite{cozzolino2019noiseprint} or employed various contrastive learning techniques to utilize consistent features~\cite{zhou2023pre,wu2023rethinking,niloy2023cfl}. 

Our method aligns with this research trend of using consistency but stands out in several key ways. Firstly, we employ a symmetric pretraining approach that focuses on source regions without distinguishing between authentic and tampered areas. Secondly, we use point prompting, which enables multi-source partitioning and addresses label agnosticity. Lastly, clustering is performed at the prediction map level, rather than on all patch pairs or individual pixels.


\subsection{Segment Anything Model}
Foundation models have first emerged in the field of Natural Language Processing (NLP), which is pretrained on large datasets and then fine-tuned across a variety of sub-tasks or domains for specific applications~\cite{bommasani2021opportunities}. 
These NLP-based foundation models have demonstrated breakthrough performance in natural language understanding and generation tasks. Their impact has expanded to other AI domains, including computer vision and speech recognition, leading to models like CLIP~\cite{radford2021learning} and wav2vec 2.0~\cite{baevski2020wav2vec}.

Recently, Meta AI introduced SAM ~\cite{kirillov2023segment} as the first foundation model for image segmentation. 
As a prompt-based model, SAM accepts point prompts, bounding boxes, and masks. Furthermore, its design allows for integration with other models to handle text prompts, enabling flexible integration with other systems.
SAM has been fine-tuned and applied to various domains such as polyp segmentation~\cite{Li2023Polyp, Zhou2023Can}, camouflaged object detection~\cite{tang2304can}, and others~\cite{ji2023segment}, with related research publications keep emerging.

\subsubsection{SAM in IFL.}
Very recently, there have been a few attempts to use SAM in IFL techniques.
One approach~\cite{su2024novel} constructs an IFL model by adding an SRM filter~\cite{zhou_learning_2018} to SAM. It completely removes the prompt encoder, essentially using SAM as a modern segmentation backbone.
Another study~\cite{karageorgiou2024fusion} fuses various signals using attention, and to assist with this attention, it employs a pretrained and frozen SAM for instance segmentation.

In summary, previous studies have primarily used SAM either as a backbone or solely to obtain segmentation masks. These approaches neglect SAM's most significant feature—its promptable capability—and fail to fully leverage its potential.
Meanwhile, we pioneer the application of promptable segmentation models for partitioning images into source regions. 
By using SAM-based point prompts, we enable each point to serve as a reference, segmenting the source region that contains it.
Moreover, inspired by SAM's automatic mask generation process, we propose an inference technique that involves placing points on the image in a grid pattern and aggregating the results. This approach enables, for the first time, multi-source partitioning.





\section{Method}
\label{sec:method}

\subsection{Overview}

The core methodology we propose, the SAFIRE framework, refers to the pretraining, training, and inference processes for IFL. 
The neural network used within this framework is termed the SAFIRE model, which does not require a specific structure and can be freely modified. 
In this paper, we utilize a slightly altered structure from SAM, adding only adapter layers to the image encoder for enhancing the model's capability to extract forensics features by utilizing low-level signals (outlined in Fig. \ref{saf.fig.main} and detailed in Appendix).
It consists of an image encoder $E(\cdot)$, prompt encoder $F(\cdot)$, and mask decoder $D(\cdot,\cdot)$. 
The model takes an image $I$ and a point prompt $P$ as inputs and outputs a prediction map $X$ and confidence score $s$ for the source region that includes the point.

The upcoming sections will delve into a detailed explanation of the SAFIRE framework. Initially, for effective source image partitioning, the image encoder is pretrained through region-to-region contrastive learning. Subsequently, in the main training phase, the model is trained on source region segmentation using point prompts. In the final inference stage, multiple points are fed into the model in a grid formation, and all the results are aggregated to obtain the final prediction heatmap.

\subsection{Pretraining: Region-to-Region Contrastive Learning}
\label{sec:method.pretraining}

\begin{figure}[t]
\centering{\includegraphics[width=1.0\linewidth]{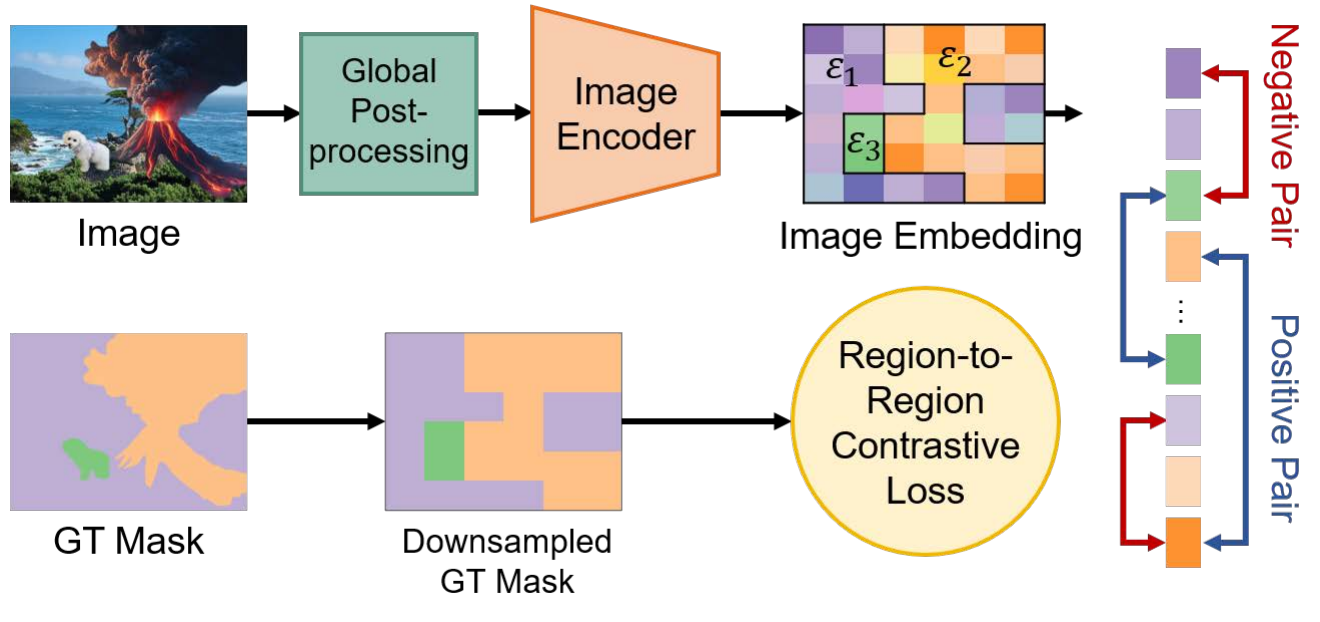}}
\caption{Pretraining. 
Features originating from the same source region become closer in the feature space, while those from different source regions move apart, enabling the image encoder to learn information that distinguishes source regions.}
\label{saf.fig.pretrain}
\end{figure}


We propose \textbf{Region-to-Region Contrastive Learning} to pretrain the image encoder for effective source region partitioning (Fig. \ref{saf.fig.pretrain}). This approach aims to have embeddings from the same source region close together in the feature space, while those from different source regions are distanced, when an image consists of two or more sources.

Leveraging the proven effectiveness of the InfoNCE loss in contrastive learning~\cite{oord2018representation}, we define our loss function as follows.
Let $I \in \mathbb{R}^{3 \times H \times W}$ be an input image composed of $r$ sources, $E(\cdot)$ the image encoder, and $\mathcal{E}=E(I) \in \mathbb{R}^{V \times \frac{H}{K} \times \frac{W}{K}}$ the image embeddings with downsampling ratio $K$. 
With a slight abuse of notation, we treat $\mathcal{E}$
as a set of $V$-dimensional image embeddings. Then there are $\frac{H}{K} \times \frac{W}{K}$ embeddings $q\in \mathbb{R}^V$ in $\mathcal{E}$. 
We also let $\{\mathcal{E}_i\}_{i=1}^r$ be the partition of $\mathcal{E}$ which corresponds to source regions in $I$. 

Then we define the region-to-region contrastive loss $\mathcal{L}_{R2R}$ as:
{\fontsize{7pt}{9pt}
\begin{align}
\label{saf.eqn.r2r}
    InfoNCE(q,p,N) &= -\textrm{log}\left(\frac{\exp\left(\frac{q\cdot p}{\tau}\right)}{\exp\left(\frac{q\cdot p}{\tau}\right)+\sum_{n \in N}\exp\left(\frac{q\cdot n}{\tau}\right)}\right),   \\
    \mathcal{L}_{R2R} &= \frac{1}{|\mathcal{E}|} \sum_{i=1}^r \sum_{q \in \mathcal{E}_i} InfoNCE\Big(q, \overline{\mathcal{E}_i \backslash \{q\}}, \mathcal{E}\backslash \mathcal{E}_i \Big),
\end{align}
}
where $\tau$ is a hyperparameter called temperature, $|\cdot|$ returns the number of elements and $\overline{\cdot}$ returns the average over all elements.

Before the image passes through the image encoder, global post-processing such as various blurring, noise addition, or contrast changes are probabilistically applied to it.
By doing so, we expect the image encoder to become robust to global common variations and focus more on fine local distinctions.

Given the large size of the image encoder, we have determined that the currently available public forgery datasets are insufficient in scale and noisy. Therefore, we generate and use a large-scale noise-free dataset, SafireMS-Auto. More in Appendix.

\subsection{Training: Source Region Segmentation Using Point Prompts}

\begin{figure}[t]
\centering{\includegraphics[width=1.0\linewidth]{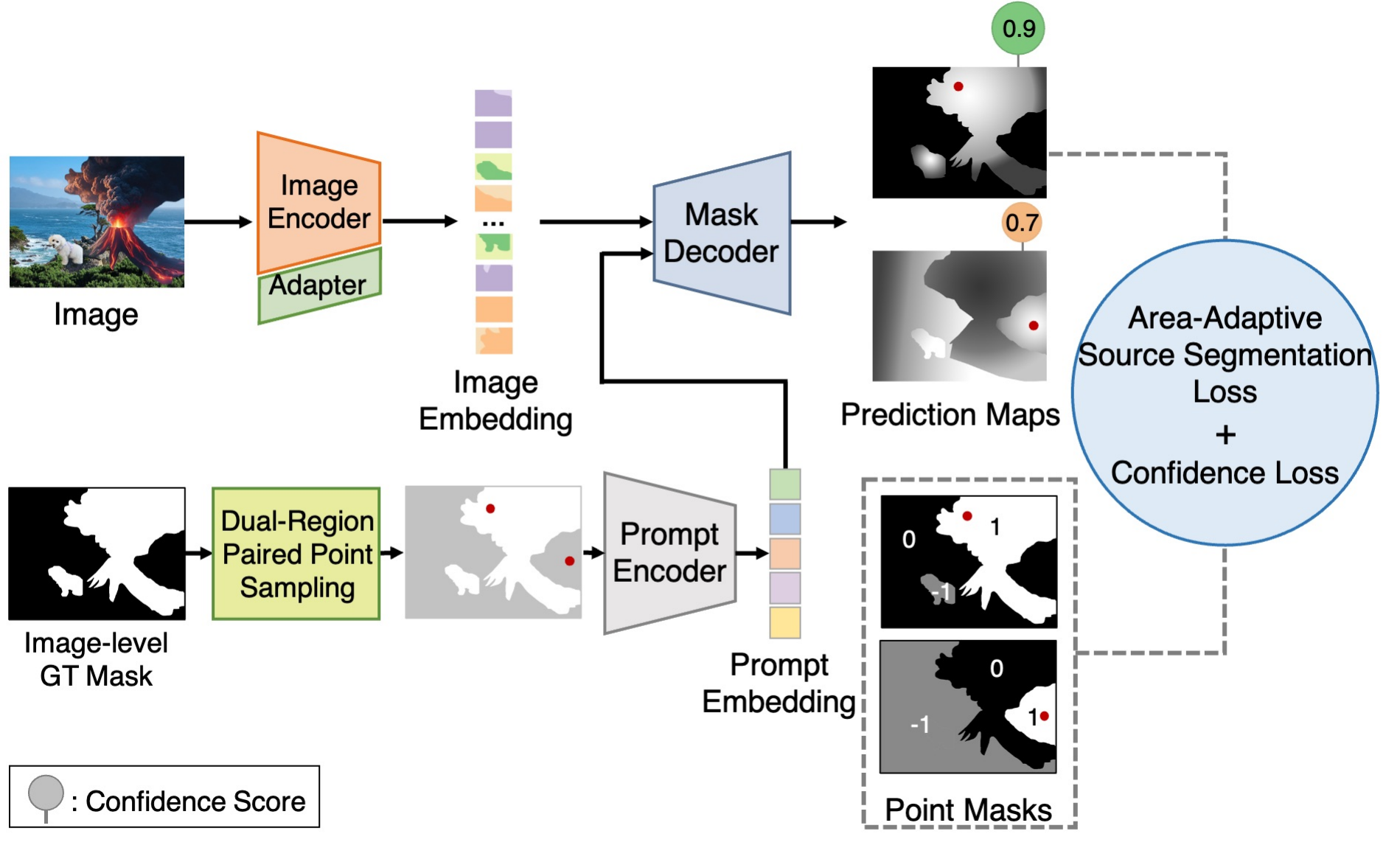}}
\caption{Training. The adapter and mask decoder are trained to segment the source region that includes the given point effectively. Furthermore, it is trained to output a confidence score of this prediction map for inference purposes.}
\label{saf.fig.train}
\end{figure}

Upon completion of the image encoder pretraining, the SAFIRE model undergoes the main training to accurately segment the source region in response to the specified point prompt (Fig. \ref{saf.fig.train}).
Both the image encoder and prompt encoder are frozen: the image encoder in its pretrained state and the prompt encoder in its original SAM state.
The adapter component and mask decoder are trained by feeding image embeddings and prompt embeddings into the mask decoder, ensuring the output aligns with the correct mask. 

\subsubsection{Point Mask Creation.}
During the training process, it is necessary to transform the image-level ground truth mask into a mask corresponding to the given point, which we call a point mask.
If there is a multi-source mask where different labels are assigned to each source region, then a point mask could be simply created by assigning 1 to the source region that contains the point and 0 otherwise.
However, almost all of the datasets currently available for IFL tasks are in only binary form, marking manipulated parts as 1 and unaltered parts as 0.

We introduce a methodology to convert these image-level binary masks into point masks. If a manipulated image uses only two sources, the areas marked as 0 and 1 would each represent a single source region. Taking a step further, we also consider connected components.
A connected region containing a given point is marked as 1, and other connected regions neighboring this region are labeled as 0. Regions that are not neighboring it are assigned an ignore label of -1, which is ignored when calculating losses.
This transformation allows us to train for multi-source partitioning using only datasets with binary labels.

To be specific, let $Y \in {\{0,1\}}^{H \times W}$ be the ground truth mask for an image $I$ which contains $c$ connected components, 
$R=\{(i,j)\in \mathbb{Z}^2:0\leq i < H, 0\leq j < W\}$ a set of integer coordinates of $I$,
$\{R_i\}_{i=1}^c$ the partition of $R$ covering connected components of $Y$,
$P \in R$ a point prompt, and $R^{P}$ a region contains $P$ which is one of $\{R_i\}_{i=1}^c$.
Then the point mask $Y_P \in {\{-1,0,1\}}^{H \times W}$ can be computed as:
\begin{align}
 \label{saf.eqn.point_mask}
 Y_P[i,j]=
 \begin{cases}
    1, & \text{if }\, (i,j) \in R^P\\
    0, & \text{if }\, (i,j) \in neighbors(R^P)\\
    -1, & \text{otherwise}
\end{cases},
\end{align}
where $neighbors(\cdot)$ returns the union of neighboring regions.

\subsubsection{Dual-Region Paired Point Sampling.}
The image encoder computes image embeddings independently of the point prompts.
Maximally leveraging this feature, efficient training can be achieved by simultaneously processing multiple point prompts for a single image.
Furthermore, to balance the source regions, points were always sampled in pairs from regions marked as 0 and 1 based on the image-level ground truth.

\subsubsection{Area-Adaptive Source Segmentation Loss.}
For each point, we can define a loss function that minimizes the difference between the prediction map and the point mask (Fig.~\ref{saf.fig.train}).
Here, not all pixels within a point mask contribute equally to the loss because doing so would result in smaller areas being overlooked. 
Traditional IFL techniques have addressed the similar issue of manipulated areas being small in most images by assigning greater weight to tampered class~\cite{kwon2022learning}. However, in our point masks, there is no distinction between manipulated and pristine regions; there exist only multiple source regions. Therefore, we use a strategy that assigns greater weight to smaller areas within each point mask, regardless of whether the correct label in those areas is 0 or 1. This differs from the class-specific weights used in most semantic segmentation tasks in that the weights are calculated within a single image~\cite{wang_deep_2020}.

Let $I$ be an input image, $P$ a point prompt, $(X,s)=D(E(I),F(P))$ the output of the mask decoder where $X$ is the prediction map and $s$ is the confidence score, and $Y_P$ the ground truth point mask for $P$. We only compute the loss within the valid label region $R^{Y_P, \{0,1\}}$ by letting $R^{A, B}=\{(i,j) \in R: A[i,j] \in B \}$.
Then the Area-Adaptive Source Segmentation Loss $\mathcal{L}_{AASS}$ is defined as:
{\fontsize{6.5pt}{7pt}
\begin{align}
\label{saf.eqn.aaxent}
    \mathcal{L}_{AASS} = & -\mathop{\mathbb{E}}_{(i,j)} \bigl[ w_1 \cdot Y[i,j] \cdot \textrm{log} \left( \sigma(X[i,j]) \right) \notag \\
    & + w_0 \cdot (1-Y[i,j]) \cdot \textrm{log} \left(1-\sigma(X[i,j])\right) \bigr], \notag
\end{align}
\begin{align}
    w_1 = \textrm{min} \left( \frac{|R^{Y_P, \{0,1\}}|}{|R^{Y_P, \{1\}}|}, C_{AASS} \right), \; \text{and} \; 
    w_0 = \textrm{min} \left( \frac{|R^{Y_P, \{0,1\}}|}{|R^{Y_P, \{0\}}|}, C_{AASS} \right),
\end{align}
}
where the expectation is calculated over $R^{Y_P, \{0,1\}}$, $\sigma(\cdot)$ is a sigmoid function, and $C_{AASS}$ is a hyperparameter to limit the weight.

\subsubsection{Confidence Loss.}
For use in inference, the mask decoder also predicts confidence scores.
Unlike SAM which predicts box-level mean intersection over union (mIoU) score to be predicted, our model predicts pixel accuracy to measure the performance globally rather than rectangular mIoU. The confidence score loss $\mathcal{L}_{conf}$ is defined as:
\begin{align}
\label{saf.eqn.confloss}
    \mathcal{L}_{conf} = \mathop{MSE}_{R^{Y_P,\{0,1\}}} \left( acc \left( bin(X), Y_P \right), s \right) ,
\end{align}
where $bin(\cdot)$ thresholds the input into binary maps, converting values greater than 0 to 1 and values less than or equal to 0 to 0, $MSE(\cdot,\cdot)$ returns the pixel-wise mean squared error, and $acc(\cdot)$ returns the accuracy.

\subsubsection{Total Training Loss.}
Finally, we obtain the total training loss $\mathcal{L}_{train}$ as follows:
\begin{align}
\label{saf.eqn.trainloss}
    \mathcal{L}_{train} = \mathcal{L}_{AASS} + \lambda_{conf} \cdot L_{conf},
\end{align}
where $ \lambda_{conf}$ is a hyperparameter balancing the two losses.

\subsection{Inference: Multiple Points Aggregation}

\begin{figure}[t]
\centering{\includegraphics[width=1.0\linewidth]{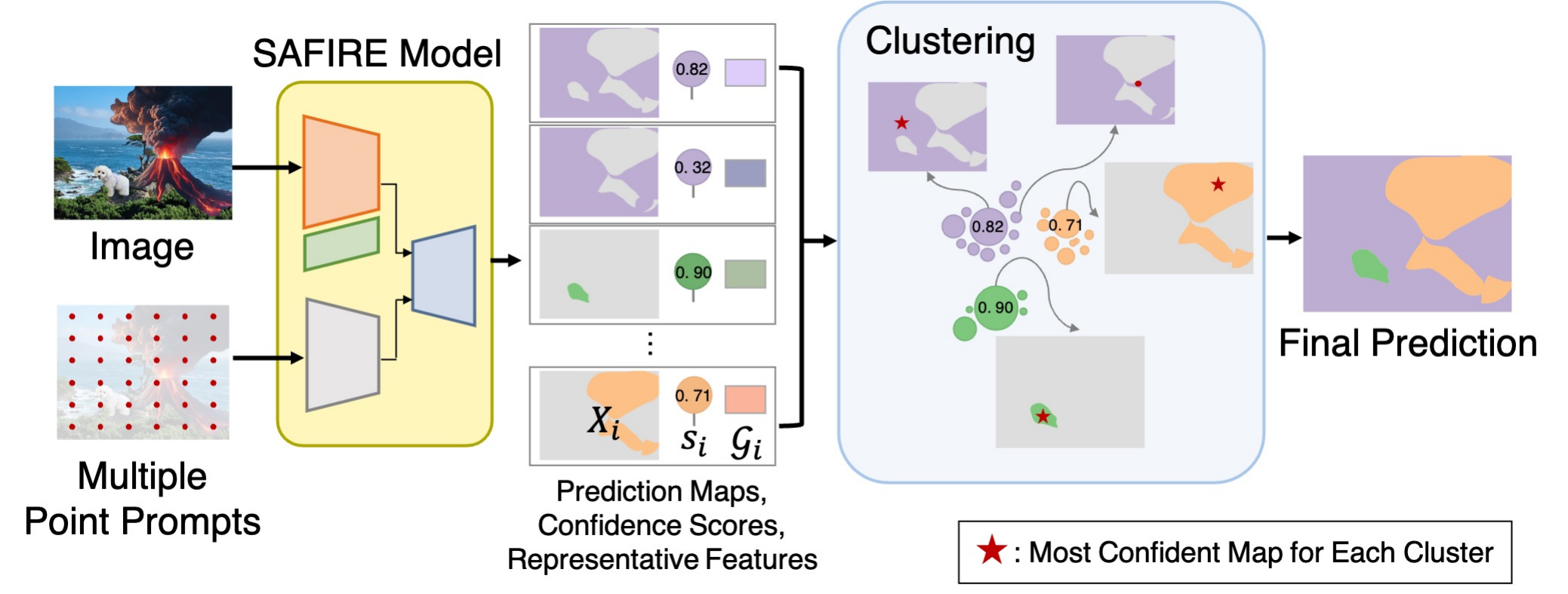}}
\caption{Inference. Multiple points in a grid pattern are input, and a prediction map is obtained for each point. Clustering is performed using the corresponding representative features, and the final prediction is produced.}
\label{saf.fig.inference}
\end{figure}

Inference is conducted using multiple point prompts (Fig. \ref{saf.fig.inference}). Alongside the image to be inferred, points are provided as input to the model in a grid format (for example, $16 \times 16$). 
The output masks are aggregated to obtain the final prediction, which could be a multi-source map or binary map.

Let $I$ be an input image, and $P_1, \cdots, P_N$ point prompts. 
First, we compute the image embedding $\mathcal{E}=E(I)$ and prompt embeddings $\mathcal{F}_i=F(P_i)$ for all $i$.
Since image embedding extraction is independent of the point prompt, it is performed only once per image. Thus, the total computation does not increase too much even if we use many points.

Thereafter, the image embedding and point embeddings pass through the mask decoder and so a prediction corresponding to each point can be obtained. 
The output of the mask decoder $D(\cdot,\cdot)$ can be expressed as: 
\begin{align}
(\{ X_1, \cdots, X_N \}, \{ s_1, \cdots, s_N \}) = D(\mathcal{E},\{\mathcal{F}_1, \cdots, \mathcal{F}_N \}),
\end{align}
where $X_i$ is a prediction map and $s_i$ is a confidence score of $X_i$.

The next step is to compute a representative feature for each prediction $X_i$, which is the average of image embeddings corresponding to the prediction area.
We define a function
$g:\mathbb{R}^{H \times W} \to \mathbb{R}^V$ by:
\begin{align}
g(X) = \frac{1}{|\mathcal{R}^{bin(\mathcal{X}),\{1\}}|}\sum_{(i,j) \in \mathcal{R}^{bin(\mathcal{X}),\{1\}}} \mathcal{E} [i,j],
\end{align}
where $\mathcal{R}$ is a set of integer coordinates of $\mathcal{E}$ and $\mathcal{X}$ is the downsampled prediction map of $X$ to match the resolution with $\mathcal{R}$.
Here, $\mathcal{R}^{bin(\mathcal{X}),\{1\}}$ represents the set of coordinates in the embedding space corresponding to the area segmented by the prediction $X$.
The representative features can be expressed as $\mathcal{G}_i = g(X_i)$ for all $i$.

Subsequently, we cluster the representative features.
The clustering is predicated on the assumption that the SAFIRE model accurately extracts features, which results in features from the same source region being gathered together.
We cluster $\{ \mathcal{G}_1, \cdots, \mathcal{G}_N \}$ into $M$ clusters $C_1, \cdots, C_M$. 
Any clustering algorithm could be applied and $M$ can be fixed in advance or regressed by the algorithm.
For general source region partitioning, we may allow the algorithm to determine the proper $M$. 
In situations where the number of sources is known, algorithms with a fixed number of clusters can be used.

Afterward, the most confident mask from each cluster is selected.
Each cluster represents one source region of the input image and the most confident mask corresponds to the best prediction of it. 
We collect indices of the maximum confidence scores for each cluster:
\begin{align}
j^* = \mathop{\text{argmax}}_{\mathcal{G}_i \in C_j} s_i.
\end{align}

Finally, these masks are combined to obtain the final prediction.
The simplest method is taking the softmax:
\begin{align}
X^* = \text{softmax}\{ X_{1^*}, \cdots, X_{M^*}\}.
\label{saf.eqn.final_pred}
\end{align}
For the special case when $M=2$, to obtain a binary prediction map, the simple average of the two predictions produces an effective output:
\begin{align}
X^* = \frac{1}{2}\{ \sigma(X_{1^*}) + (1-\sigma( X_{2^*}))\}.
\end{align}


\section{Experiments on binary IFL}
\label{saf.sec:exp_ifl}
We begin with the traditional task of localizing forged regions in images. 
Note that SAFIRE can make binary predictions as well as multi-source predictions.

\subsection{Experimental Settings}
\subsubsection{Implementation Details.}
Our model undergoes pretraining followed by training. 
The temperature $\tau$ for the region-to-region contrastive learning in Eq. \eqref{saf.eqn.r2r} is set to $0.1$. 
The weight limit $C_{AASS}$ for the AASS loss in Eq. \eqref{saf.eqn.aaxent} is set to $10$ and $\lambda_{conf}=0.1$ in Eq. \eqref{saf.eqn.trainloss}.
During the inference phase, $M$ is fixed to 2 to obtain predictions in binary form.
We use $16 \times 16$ point prompts and k-means clustering.

\subsubsection{Datasets.}
We train the network using a commonly adopted setting~\cite{guillaro2023trufor} that incorporates four datasets ~\cite{kniaz_point_2019,novozamsky_imd2020_2020,dong_casia_2013,kwon2022learning} which consists of real and fake images, also known as the CAT-Net~\cite{kwon2022learning} setting. We test the performance using five public datasets which have no overlap with the training datasets: Columbia~\cite{ng_data_2004}, COVERAGE~\cite{wen2016coverage}, CocoGlide~\cite{guillaro2023trufor}, RealisticTampering~\cite{korus2016multi}, and NC16~\cite{guan_mfc_2019}. These consist of various forgery types including splicing, copy-move, removal, and adding objects using generative models. 
During testing, images are input in their original form, except for the NC16 dataset, where images were scaled down due to memory constraints in some comparative methods.

\renewcommand{\arraystretch}{1.0}
\newcolumntype{M}{>{\centering\arraybackslash}m{0.645cm}}

\begin{table}[t]
\centering
\begin{subtable}[t]{\columnwidth}
{\fontsize{9pt}{10pt}\selectfont
\begin{tabularx}{\columnwidth}{@{}lMMMMMM@{}}
\toprule
\multicolumn{1}{c}{Method} & Col. & COV. & CG. & RT. & NC16 & Avg. \\ \midrule
EXIF-SC       & 78.8          & 16.2          & 29.6          & 14.4          & 16.8          & 31.2          \\
ManTraNet     & 50.5        & 31.2                        & 51.6                         & 21.5                          & 19.9                    & 35.0                    \\
SPAN          & 39.7                        & 16.1                        & 29.6                         & 8.7                           & 11.2                    & 21.1                    \\
AdaCFA        & 58.4                        & 17.9                        & 28.7                         & 22.3                          & 11.4                    & 27.7                    \\
CAT-Net v2    & \underline{85.8}                  & 37.6                        & 43.3                         & 14.3                          & 28.2                    & 41.8                    \\
IF-OSN        & 74.7              & 29.9                        & 42.8                         & 33.4                          & 32.5                    & 42.7                    \\
MVSS-Net      & 72.7                        & 50.8                        & 48.6                         & 17.5                          & 32.7                    & 44.5                    \\
PSCC-Net      & 60.0                        & 46.6                        & 51.7                         & 9.7                           & 13.4                    & 36.3                    \\
TruFor        & 85.7                        & \underline{58.7}                  & \underline{52.2}                   & \textbf{43.2}                 & \underline{41.6}              & \underline{56.3}              \\
NCL           & 47.3	                     & 21.3                        & 35.8                         & 14.5	                          & -                   & -                    \\
SAM           & 40.0	                     & 18.1                       & 33.9                         & 8.1	                          & 11.2                    & 22.3                    \\
SAFIRE (Ours) & \textbf{97.9}               & \textbf{63.4}               & \textbf{63.5}                & \underline{39.3}                    & \textbf{48.8}           & \textbf{62.6}           \\ \bottomrule
\end{tabularx}}
\end{subtable}

\vspace{2mm}

\begin{subtable}[t]{\columnwidth}
{\fontsize{9pt}{10pt}\selectfont
\begin{tabularx}{\columnwidth}{@{}lMMMMMM@{}}
\toprule
\multicolumn{1}{c}{Method} &  Col. & COV. & CG. & RT. & NC16 & Avg. \\ \midrule
EXIF-SC       & \underline{92.9}    & 31.7          & 42.0          & 28.6          & 33.6          & 45.8          \\
ManTraNet     & 64.6          & 47.9          & 67.3          & 26.4          & 27.7          & 46.8          \\
SPAN          & 49.9          & 24.8          & 38.1          & 15.6          & 16.8          & 29.0          \\
AdaCFA        & 62.7          & 21.5          & 36.3          & 24.1          & 14.0          & 31.7          \\
CAT-Net v2    & 92.1          & 57.5          & 60.3          & 24.3          & 43.4          & 55.5          \\
IF-OSN        & 82.9          & 46.4          & 59.1          & 46.6          & 45.6          & 56.1          \\
MVSS-Net      & 78.1          & 64.6          & 64.2          & 29.4          & 44.0          & 56.0          \\
PSCC-Net      & 75.7          & 60.6          & 68.5          & 18.3          & 28.4          & 50.3          \\
TruFor        & 91.3          & \underline{72.1}    & \underline{72.0}    & \textbf{53.3} & \underline{54.7}          & \underline{68.7}    \\
NCL & 62.5          & 32.4          & 49.7          & 27.0          & - & -          \\
SAM           & 47.8          & 35.7          & 46.3          & 17.5          & 23.3          & 34.1          \\
SAFIRE (Ours) & \textbf{99.7} & \textbf{76.9} & \textbf{76.9} & \underline{49.9}    & \textbf{61.4}    & \textbf{73.0}     \\ \bottomrule
\end{tabularx}}
\end{subtable}
\caption{IFL results using F1 fixed (\%, top) and F1 best (\%, bottom). Under both metrics, SAFIRE achieves the highest forgery localization performance in four out of five datasets and is ranked first in terms of average score.}
\label{saf.tab.f1_both}
\end{table}

\subsubsection{Comparative Methods.}

Following the protocol from~\cite{guillaro2023trufor}, we ensure a fair comparison by selecting recent techniques with publicly accessible code and pretrained models, trained on disjoint datasets from test sets.
Namely, these include ManTra-Net~\cite{wu_mantra-net_2019}, SPAN~\cite{hu2020span}, AdaCFA~\cite{bammey2020adaptive}, CAT-Net v2~\cite{kwon2022learning}, IF-OSN~\cite{wu2022robust}, MVSS-Net~\cite{dong2022mvss}, PSCC-Net~\cite{liu2022pscc}, TruFor~\cite{guillaro2023trufor}, and NCL~\cite{zhou2023pre}.
Furthermore, we also utilize a pure SAM~\cite{kirillov2023segment} model trained on the same datasets.


\subsubsection{Metrics.}
We evaluate the localization performance in the same manner as conducted in the TruFor~\cite{guillaro2023trufor} paper. Specifically, performances are reported in terms of permuted F1 score~\cite{huh_fighting_2018, kwon2022learning} using both fixed 0.5 threshold (F1 fixed) and best threshold per image (F1 best).

\subsection{Evaluation Results}
\label{saf.subsec.IFL_results}

Table \ref{saf.tab.f1_both} shows a comparative analysis of binary IFL performance. A hyphen (`-') indicates that the dataset was used for training, and so excluded. Notably, SAFIRE shows superior performance on both F1 fixed and F1 best, achieving the first rank in four out of five datasets. 
Moreover, the average score on all datasets reaffirms the superiority of SAFIRE, securing the best position in overall performance. 
In addition, in the Appendix, we can see that our method also outperforms other techniques under various global post-processing conditions, proving its robustness.

\begin{figure}[]
\centering{\includegraphics[width=1.0\linewidth]{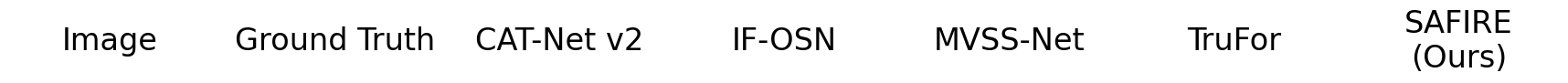}}
\centering{\includegraphics[width=1.0\linewidth]{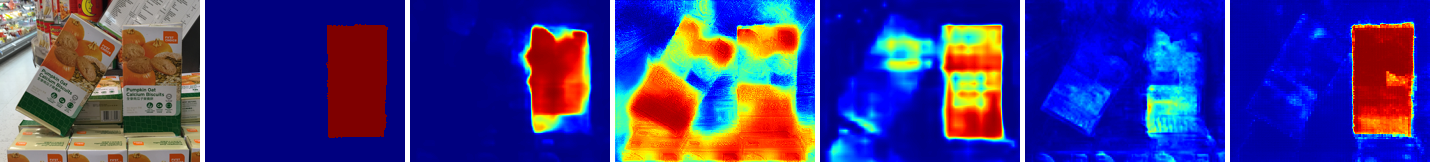}}
\centering{\includegraphics[width=1.0\linewidth]{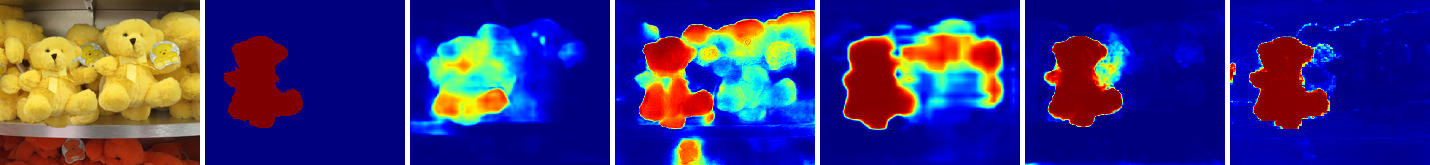}}
\centering{\includegraphics[width=1.0\linewidth]{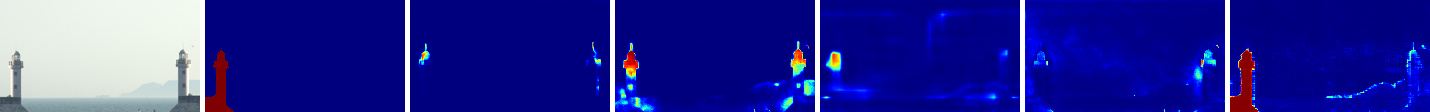}}
\centering{\includegraphics[width=1.0\linewidth]{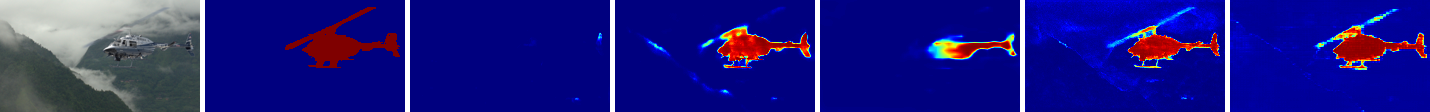}}
\centering{\includegraphics[width=1.0\linewidth]{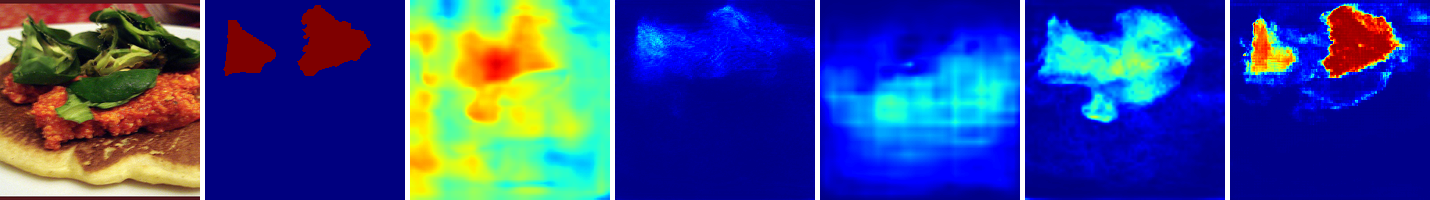}}
\centering{\includegraphics[width=1.0\linewidth]{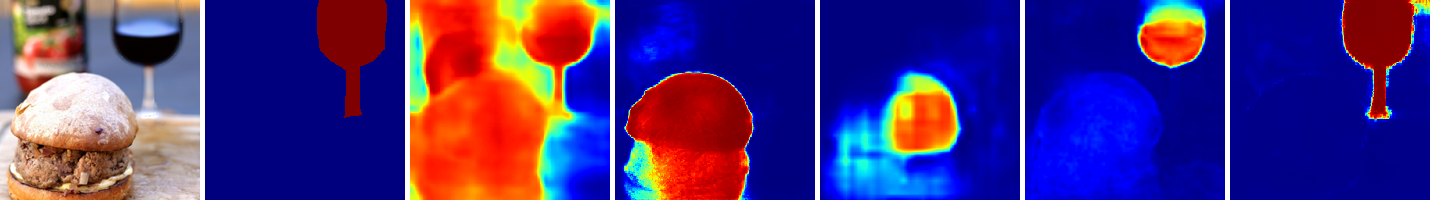}}
\centering{\includegraphics[width=1.0\linewidth]{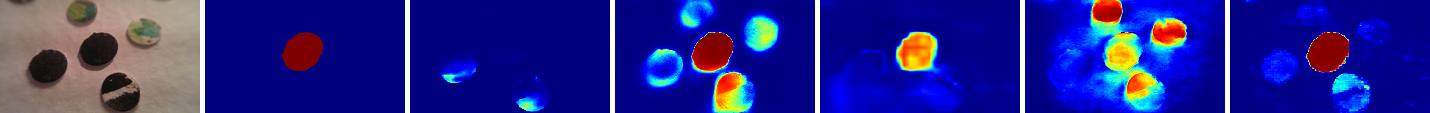}}
\centering{\includegraphics[width=1.0\linewidth]{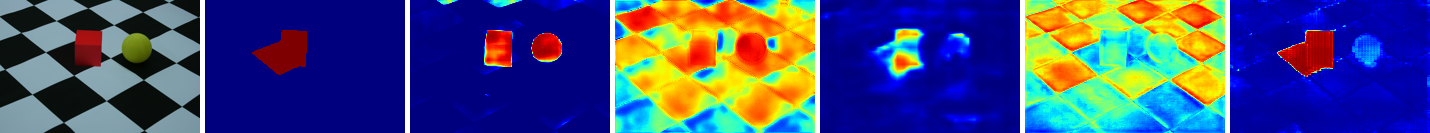}}
\centering{\includegraphics[width=1.0\linewidth]{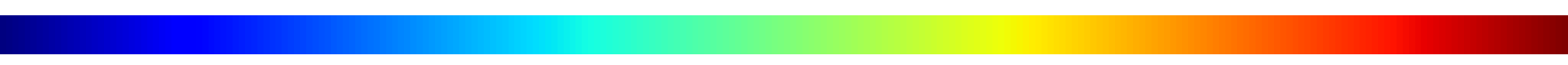}}
\caption{Visualization of IFL. The colors indicate the confidence of forged pixels.}
\label{saf.fig.ifl}
\end{figure}

Figure \ref{saf.fig.ifl} shows the qualitative results of IFL produced by each model. 
SAFIRE successfully identifies sophisticated and challenging manipulations that other techniques fail to detect, with less false positive detection.
In particular, SAFIRE achieves significantly more accurate predictions in sophisticated AI-generated partial manipulations compared to other techniques.

\subsubsection{Ablation Study.}

\newcolumntype{M}{>{\centering\arraybackslash}m{0.47cm}}
\begin{table}[t]
{\fontsize{9pt}{10pt}\selectfont
\begin{tabularx}{\columnwidth}{@{}l@{\hspace{1mm}}MMMMMM@{}}
\toprule
\multicolumn{1}{c}{Setting}            & R2R Loss & AASS Loss & Prom-pting & Conf. Loss & F1 fixed       & F1 best        \\ \hline
SAFIRE              & \checkmark   & \checkmark      & \checkmark  & \checkmark & \textbf{62.6} & \textbf{73.0} \\
R2R $\rightarrow$ SAM pretraining     &  -  & \checkmark      & \checkmark  & \checkmark & 48.6          & 61.0          \\
No area adaptation           & \checkmark   &    -   & \checkmark & \checkmark  & 56.9           & 71.5         \\
Prompting $\rightarrow$ Binary seg. & \checkmark   & \checkmark      &  -  & - & 28.0          & 37.4          \\ 
Using random mask & \checkmark   & \checkmark        & \checkmark &  - & 43.5         & 54.5         \\ 
Baseline                & - &  - & -  & - & 22.3         & 34.1          \\\bottomrule
\end{tabularx}}
\caption{Ablation study (\%). 
All core components of SAFIRE contribute to the performance of IFL. The usage of prompting is found to be critical to performance.}
\label{saf.tab.ablation_study}
\end{table}

To ensure the completeness of our study, we conduct an ablation study on the key components of our framework: Region-to-Region Contrastive Loss, area adaptive feature in Area-Adaptive Source Segmentation Loss, point prompting, and Confidence Loss (Table \ref{saf.tab.ablation_study}). 
We substitute each with a conventional counterpart for comparison. 

The results demonstrate diminished performance in the absence of any single component compared to the full SAFIRE framework, which integrates all four. Furthermore, a baseline model excluding all four key features exhibits significantly inferior results, underscoring the indispensable role of these four components in SAFIRE.

Especially, we observe that source region partitioning based on prompting outperforms binary segmentation. 
A model with the same structure and pretraining achieves only a 28.0\% F1 fixed score when using binary segmentation. However, when employing prompting for source partitioning, the model's performance significantly improves, reaching 62.6\%.
This demonstrates the effectiveness of SAFIRE's prompting approach in enabling the network to understand the characteristics of the same source regions, resulting in stable learning and outstanding performance.

\section{Experiments on Multi-source Partitioning}
\label{sec:exp_ms}


\newcommand{\gray}[1]{\textcolor[gray]{0.7}{#1}}

\newcolumntype{M}{>{\centering\arraybackslash}m{0.53cm}}
\begin{table}[]
\centering
{\fontsize{9pt}{10pt}\selectfont
\begin{tabularx}{\columnwidth}{@{}lMMMMMM@{}}
\toprule
\multicolumn{1}{c}{\multirow{2}{*}{Method}} & \multicolumn{3}{c}{p\_mIoU (\%)} & \multicolumn{3}{c}{ARI (\%)} \\ 
                        & 2src & 3src & 4src & 2src & 3src & 4src \\
                        \cmidrule(r){1-1}
                        \cmidrule(lr){2-4} \cmidrule(lr){5-7}
CAT-Net v2         &  53.8  & - & - &  25.1  & - & - \\
IF-OSN             &  54.8  & - & - &  25.2  & - & - \\
MVSS-Net           &  51.9  & - & - &  22.5  & - & - \\
TruFor             &  82.5  & - & - &  72.3  & - & - \\
SAFIRE (k-means)    &  \textbf{90.3}  & \textbf{62.2} & \textbf{54.0} & \textbf{80.7}  & 57.8 & 55.2 \\
SAFIRE (DBSCAN)     &  89.7  & 57.6 & 48.9  &  80.2  & \textbf{60.2} & \textbf{59.4} \\\bottomrule
\end{tabularx}}
\caption{Multi-source partitioning results of SafireMS-Expert. SAFIRE can accurately partition images into multiple sources based on their origins.}
\label{saf.tab.ms_mIoU_ARI}
\end{table}

One of the unique advantages of our method is its ability to partition images composed of three or more sources into each source.
To show this capability, we conduct additional experiments on multi-source partitioning.

\subsection{Experimental Settings}
\subsubsection{Implementation Details.}
We use the same model as used in binary IFL without fine-tuning. We consider two settings for inference: the number of sources is given in advance and it is determined by the method. For the former, we use k-means clustering as done in binary IFL. For the latter, we utilize DBSCAN which automatically chooses the number of clusters in the data distribution.

\subsubsection{Datasets.}
Given that traditional forgery datasets lack instances labeled with multiple source regions, we manually constructed a multi-source dataset, dubbed SafireMS-Expert, to assess our framework's effectiveness in handling such scenarios.
The forgery types include splicing, removal by AI-based inpainting~\cite{yu2023inpaint}, reconstructing some objects using generative models, and adding objects by generative models using text prompts~\cite{zhang2023adding}. More in Appendix.


\subsubsection{Metrics.}
We propose to use permuted mIoU (p\_mIoU) and Adjusted Rand Index (ARI) for source region partitioning. We generalize p\_mIoU defined on \cite{huh_fighting_2018} for an arbitrary number of source regions. More in Appendix.

\subsection{Evaluation Results}

\begin{figure}[]
\centering{\includegraphics[width=1.0\linewidth]{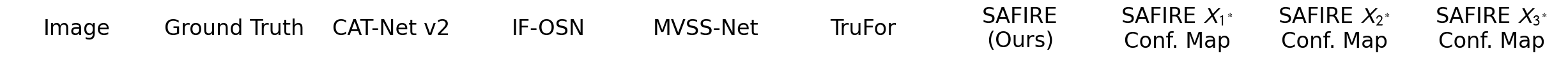}}
\centering{\includegraphics[width=1.0\linewidth]{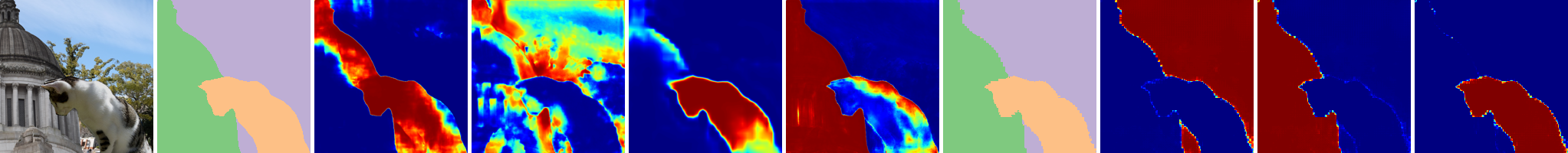}}
\centering{\includegraphics[width=1.0\linewidth]{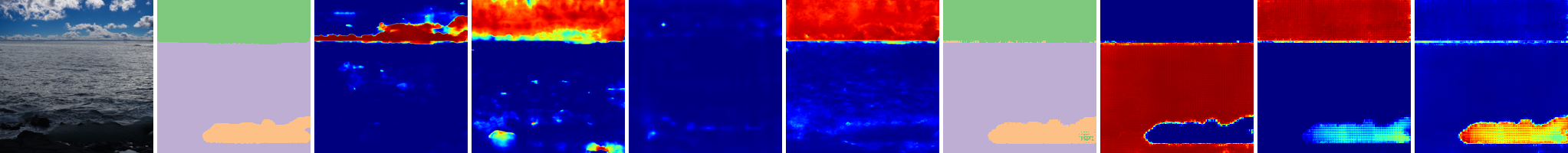}}
\centering{\includegraphics[width=1.0\linewidth]{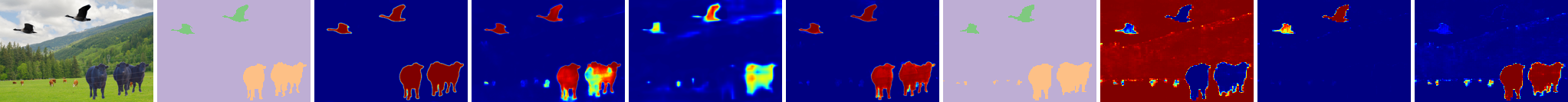}}
\centering{\includegraphics[width=1.0\linewidth]{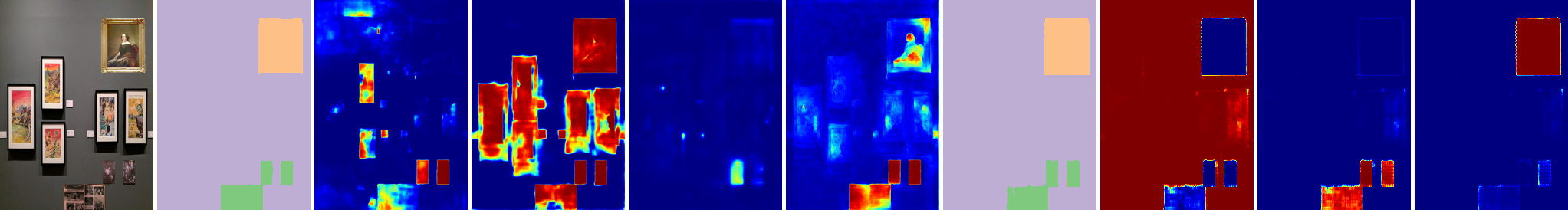}}
\centering{\includegraphics[width=1.0\linewidth]{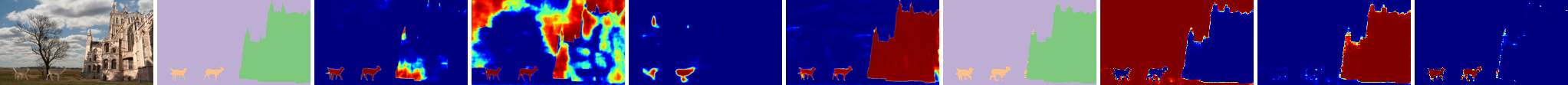}}
\centering{\includegraphics[width=1.0\linewidth]{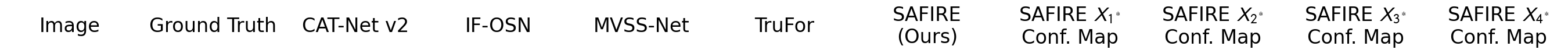}}
\centering{\includegraphics[width=1.0\linewidth]{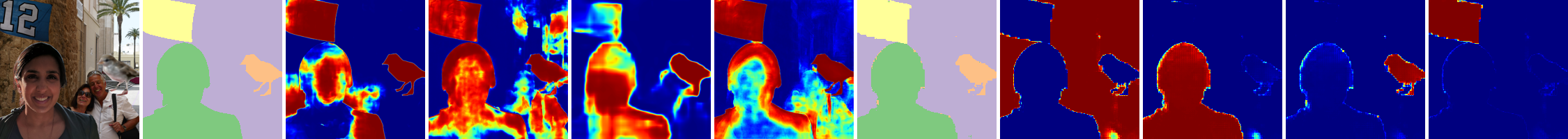}}
\centering{\includegraphics[width=1.0\linewidth]{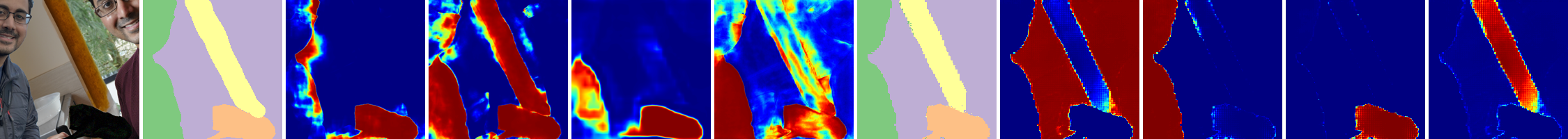}}
\centering{\includegraphics[width=1.0\linewidth]{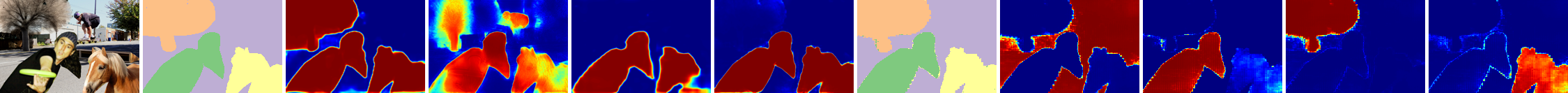}}
\centering{\includegraphics[width=1.0\linewidth]{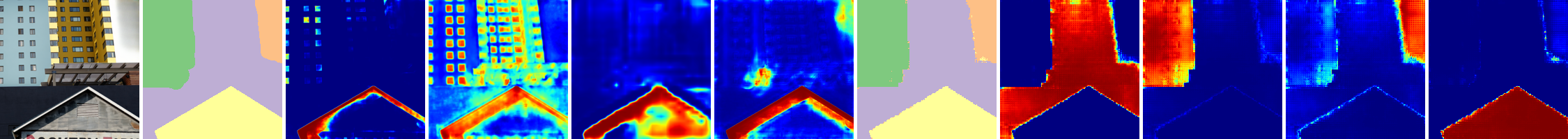}}
\centering{\includegraphics[width=1.0\linewidth]{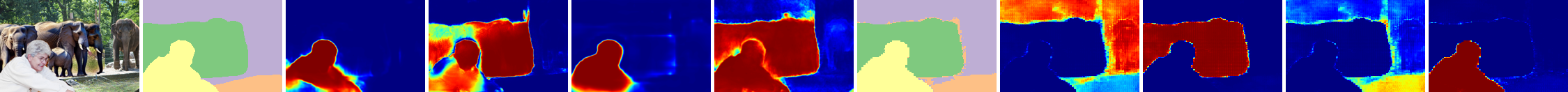}}
\caption{Visualization of multi-source IFL. Each color represents a single source region. Methods other than SAFIRE could only produce binary results.}
\label{saf.fig.ms_ifl}
\end{figure}

Table \ref{saf.tab.ms_mIoU_ARI} and Figure \ref{saf.fig.ms_ifl} present the quantitative and qualitative results of multi-source partitioning, respectively. While existing methods based on binary segmentation cannot split images into three or more sources (marked with `-'), SAFIRE can do it, even though it is trained using only binary datasets. These visualizations offer a better interpretation of forged images for humans, as manipulation often occurs multiple times in real-world scenarios. Additionally, SAFIRE’s ability to estimate the number of sources is a valuable feature. More results are in the Appendix.


\section{Conclusion}
\label{sec:concl}
Moving beyond the conventional approach of viewing the IFL tasks through binary segmentation, SAFIRE resolved this issue by partitioning images into multiple originating regions. Through region-to-region contrastive pretraining, we guided the encoder to effectively embed subtle signals for source partitioning. We utilized point prompt-based segmentation to train the SAFIRE model to accurately predict the source region containing each point. During inference, we provided point prompts in a grid format and aggregated the outputs to obtain the final prediction. Through comprehensive evaluation, SAFIRE successfully accommodated label agnosticity issues in IFL and outperformed other state-of-the-art methods. It also opened up possibilities for using point prompting in image partitioning and presented a new challenge of partitioning images into multiple source regions. It aids in comprehending the structure of the forged image and facilitates further analysis. We hope our study contributes to solving the increasingly complex image forgery issues in the era of AI.


\vspace{10mm}
\begin{appendix}
\appendix

\begin{center}
    \huge \textbf{Appendix}
    \vspace{2mm}
\end{center}

\addcontentsline{toc}{section}{Appendices}

\begin{figure*}[]
\centering{\includegraphics[width=1.0\linewidth]{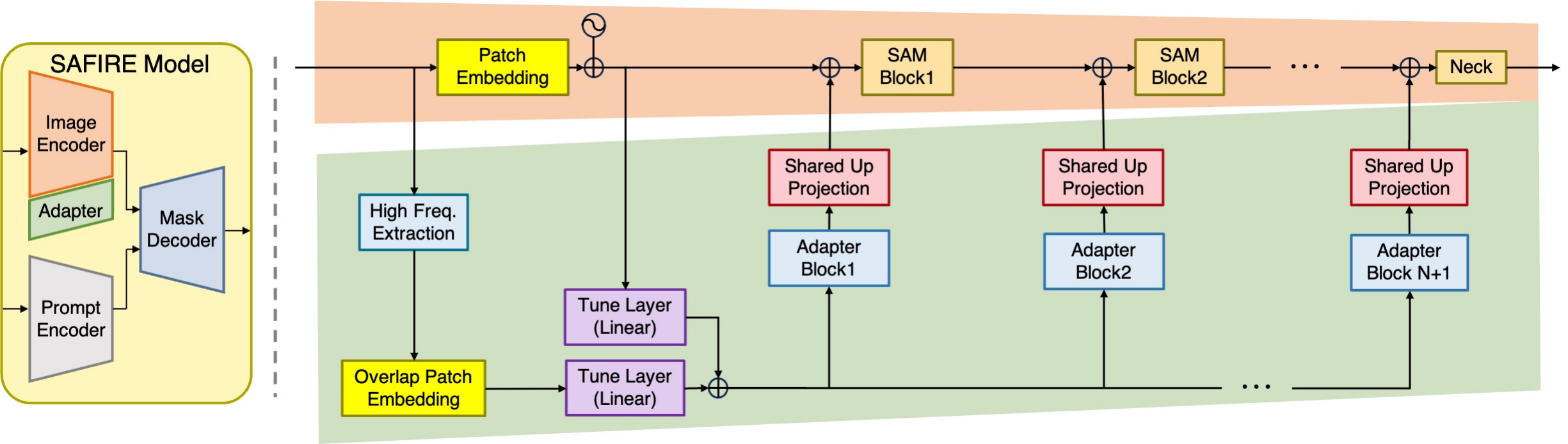}}
\caption{SAFIRE model architecture. The outline of SAFIRE model is shown on the left-hand side. The image encoder and adapter are shown on the right-hand side. 
The prompt encoder and mask decoder are the same as those in the original SAM.}
\label{saf.fig.arch}
\end{figure*}

The appendix provides supplementary details that are not included in the main manuscript due to space limitations. 
The organization of this appendix is as follows.
We begin by examining the SAFIRE model architecture. 
Subsequently, we introduce SafireMS, a new multi-source forgery dataset we have created. 
We then present the prediction maps before clustering to enhance comprehension of the SAFIRE framework. 
Next, we discuss the robustness tests to evaluate the model's stability under various global image processing and see the effect of the number of points. 
We also show more visualization results, followed by experimental environment.
Lastly, we elucidate the metrics used for multi-source partitioning.

\section{SAFIRE Model Architecture}
\label{sec:architecture}

The left-hand side of Fig.~\ref{saf.fig.arch} shows the SAFIRE model architecture as described in the main text. On the right-hand side, the structure showcases the image encoder integrated with the adapter. This design integrates adapter components into the SAM image encoder~\cite{kirillov2023segment} to strengthen the model's ability to extract forensic features. This strategy draws inspiration from recent studies that integrate adapter modules to tailor large foundational models to specific tasks~\cite{chen2023sam,houlsby2019parameter,li2022exploring,chen2022vision}.

The top section, highlighted in orange, mirrors the architecture of the SAM image encoder. 
The bottom section, depicted in green, consists of adapter components designed to refine the encoder's performance. In the context of forensic analysis, the model leverages high-frequency component extraction~\cite{liu2023explicit}.
Firstly, a fast Fourier transform (FFT) is applied and then low frequencies are removed, followed by an inverse FFT to revert it back to the image domain. 
The model processes and combines the patch embeddings from both the original and high-frequency images. Subsequently, these embeddings are fine-tuned through adapter blocks, which are linear layers, and then proceed through shared up-projection layers, fine-tuning the primary pathway.

\section{Custom Datasets: SafireMS}


This section elaborates on the creation of the SafireMS-Auto and SafireMS-Expert datasets, pivotal for the pretraining and multi-source partitioning experiments. 
We detail the generation of SafireMS-Auto, which comprises around 123k automatically generated forgery images, facilitating effective pretraining. 
We then describe the development of SafireMS-Expert, a dataset manually created by experts to include images with two to four source regions.




\renewcommand{\arraystretch}{1.2}
\newcolumntype{M}{>{\centering\arraybackslash}m{1.35cm}}
\begin{table}[]
\centering
\fontsize{9pt}{10pt}\selectfont
\begin{tabular}{@{}MMMMM@{}}
\toprule
Category            & Post-Processing & Images & Subtotal Images         & Total Images              \\ \hline
\multirow{4}{*}{CM} & Both            & 5,000  & \multirow{4}{*}{20,000} & \multirow{12}{*}{123,164} \\
                    & Background      & 5,000  &                         &                           \\
                    & Foreground      & 5,000  &                         &                           \\
                    & Neither         & 5,000  &                         &                           \\ \cline{1-4}
\multirow{4}{*}{SP} & Both            & 5,000  & \multirow{4}{*}{20,000} &                           \\
                    & Background      & 5,000  &                         &                           \\
                    & Foreground      & 5,000  &                         &                           \\
                    & Neither         & 5,000  &                         &                           \\ \cline{1-4}
\multirow{2}{*}{GN} & Yes             & 20,791 & \multirow{2}{*}{41,582} &                           \\
                    & No              & 20,791 &                         &                           \\ \cline{1-4}
\multirow{2}{*}{RM} & Yes             & 20,791 & \multirow{2}{*}{41,582} &                           \\
                    & No              & 20,791 &                         &                           \\ \bottomrule
\end{tabular}
\caption{The composition of SafireMS-Auto. It consists of four categories based on the forgery techniques, and each category is further subdivided by the usage of post-processing.}
\label{saf.tab.SafireMS-Auto}
\end{table}

\begin{figure}[]
    \centering
    \begin{subfigure}[b]{0.49\linewidth}
        \centering
        \includegraphics[width=\linewidth]{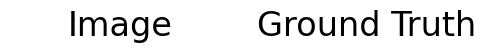}
        \includegraphics[width=\linewidth, height=1.46cm]{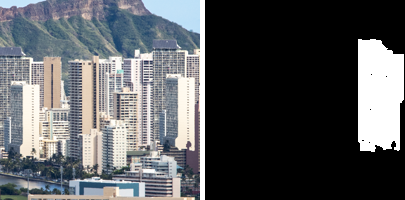}
        \includegraphics[width=\linewidth, height=1.46cm]{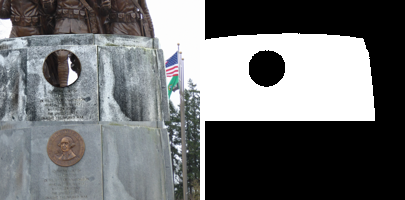}
        \includegraphics[width=\linewidth, height=1.46cm]{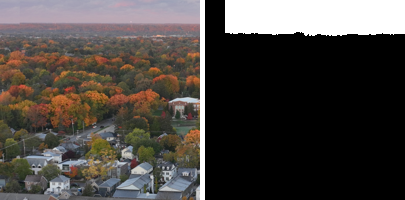}
        \caption{CM}
    \end{subfigure}
    \hfill 
    \begin{subfigure}[b]{0.49\linewidth}
        \centering
        \includegraphics[width=\linewidth]{saf_appendix_figs/text_image_SafireMS.png}
        \includegraphics[width=\linewidth, height=1.46cm]{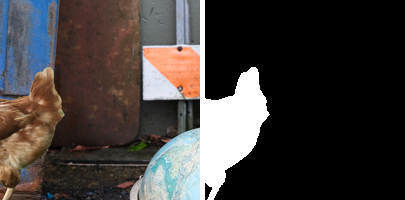}
        \includegraphics[width=\linewidth, height=1.46cm]{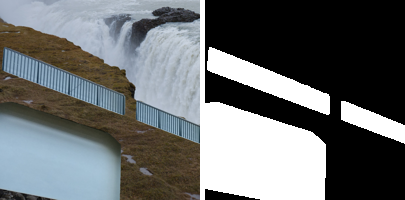}
        \includegraphics[width=\linewidth, height=1.46cm]{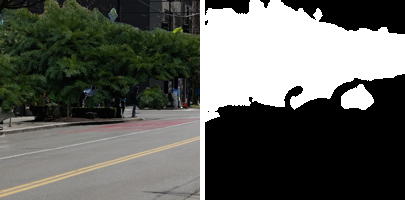}
        \caption{SP}
    \end{subfigure}
    \hfill 
    \begin{subfigure}[b]{0.49\linewidth}
        \centering
        \includegraphics[width=\linewidth]{saf_appendix_figs/text_image_SafireMS.png}
        \includegraphics[width=\linewidth, height=1.46cm]{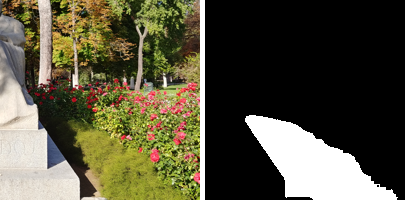}
        \includegraphics[width=\linewidth, height=1.46cm]{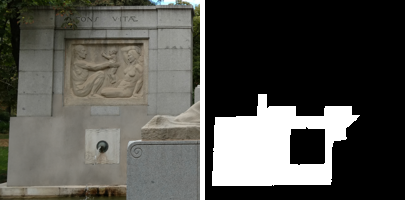}
        \includegraphics[width=\linewidth, height=1.46cm]{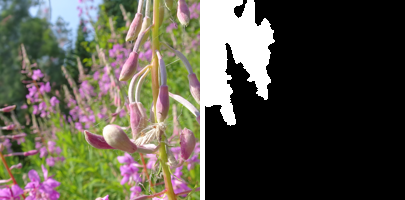}
        \caption{GN}
    \end{subfigure}
    \hfill 
    \begin{subfigure}[b]{0.49\linewidth}
        \centering
        \includegraphics[width=\linewidth]{saf_appendix_figs/text_image_SafireMS.png}
        \includegraphics[width=\linewidth, height=1.46cm]{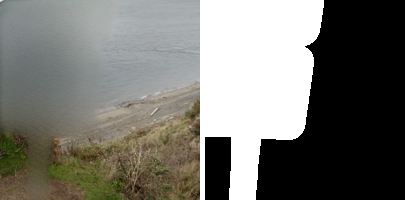}
        \includegraphics[width=\linewidth, height=1.46cm]{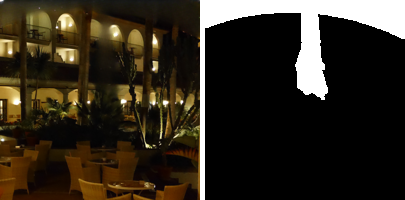}
        \includegraphics[width=\linewidth, height=1.46cm]{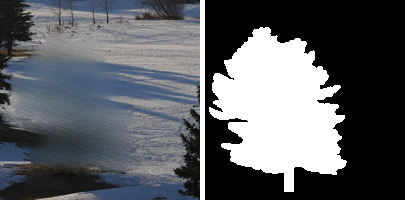}
        \caption{RM}
    \end{subfigure}
    \caption{SafireMS-Auto examples. The types of post-processing, areas for manipulation, and generation parameters are automatically selected.}
    \label{saf.fig.ms-auto}%
\end{figure}

\subsection{SafireMS-Auto}

To obtain pristine images directly from cameras without any traces of manipulation, we collect photographs from DPReview (\textnormal{www.dpreview.com}) which has lots of raw images. From the Sample Images tab on DPReview, a total of 30,244 authentic photographs are collected, with the categories set to cameras, drones, and mobiles.

Utilizing the automatic mask generation capability of SAM, multiple semantic regions are generated to partition the entire image. We select some adjacent regions randomly and union them to create a mask for the image. To ensure that the mask does not occupy an excessive area of the image, we randomly choose a threshold per image to restrict the mask area. 
Consequently, this process yields a total of 20,791 image-mask pairs.

Using these image-mask pairs, manipulated images were automatically generated.
Table \ref{saf.tab.SafireMS-Auto} provides a detailed composition. It comprises four types of forgeries: copy-move (CM), splicing (SP), generative reconstruction (GN)~\cite{zhang2023adding}, and removal by AI-based inpainting (RM)~\cite{yu2023inpaint}. These categories are meticulously curated to represent various forgery methodologies. 
The post-processing includes resizing, blurring, adding noise, decreasing resolution, and changing brightness, gamma, HSV, or contrast.
Parameters for GN and RM are chosen randomly in the predefined range.
Consequently, SafireMS-Auto consists of a comprehensive collection of 123,164 image-mask pairs, offering a diverse and extensive resource for image forensics. Figure \ref{saf.fig.ms-auto} depicts some examples.

\subsection{SafireMS-Expert}

\begin{figure}[t]
    \centering
    
    \begin{subfigure}[b]{0.32\linewidth}
        \centering
        \includegraphics[width=\linewidth]{saf_appendix_figs/text_image_SafireMS.png}
    \end{subfigure}
    \hfill 
    \begin{subfigure}[b]{0.32\linewidth}
        \centering
        \includegraphics[width=\linewidth]{saf_appendix_figs/text_image_SafireMS.png}
    \end{subfigure}
    \hfill 
    \begin{subfigure}[b]{0.32\linewidth}
        \centering
        \includegraphics[width=\linewidth]{saf_appendix_figs/text_image_SafireMS.png} 
    \end{subfigure}
    \vspace{0cm}
    
    \begin{subfigure}[b]{0.32\linewidth}
        \centering
        \includegraphics[width=\linewidth, height=1.46cm]{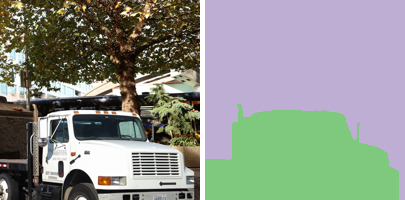}
    \end{subfigure}
    \hfill 
    \begin{subfigure}[b]{0.32\linewidth}
        \centering
        \includegraphics[width=\linewidth, height=1.46cm]{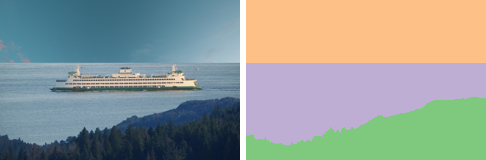}
    \end{subfigure}
    \hfill 
    \begin{subfigure}[b]{0.32\linewidth}
        \centering
        \includegraphics[width=\linewidth, height=1.46cm]{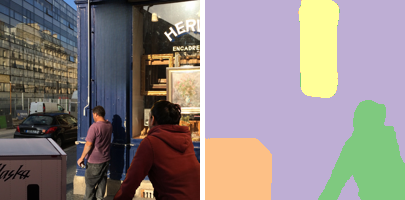} 
    \end{subfigure}
    \vspace{0cm}
    
    \begin{subfigure}[b]{0.32\linewidth}
        \centering
        \includegraphics[width=\linewidth, height=1.46cm]{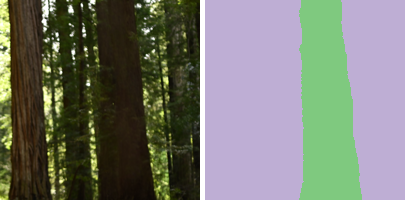}
    \end{subfigure}
    \hfill 
    \begin{subfigure}[b]{0.32\linewidth}
        \centering
        \includegraphics[width=\linewidth, height=1.46cm]{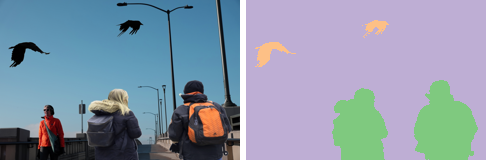}
    \end{subfigure}
    \hfill 
    \begin{subfigure}[b]{0.32\linewidth}
        \centering
        \includegraphics[width=\linewidth, height=1.46cm]{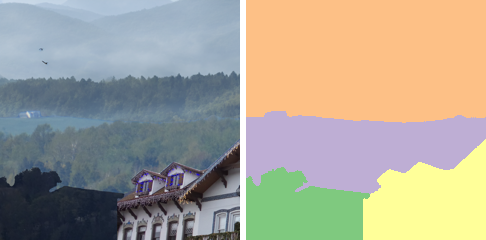} 
    \end{subfigure}
    \vspace{0cm}
    
    \begin{subfigure}[b]{0.32\linewidth}
        \centering
        \includegraphics[width=\linewidth, height=1.46cm]{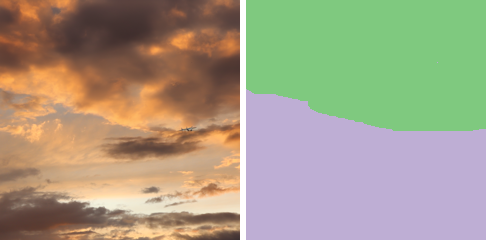}
        \caption{Two-source}
    \end{subfigure}
    \hfill 
    \begin{subfigure}[b]{0.32\linewidth}
        \centering
        \includegraphics[width=\linewidth, height=1.46cm]{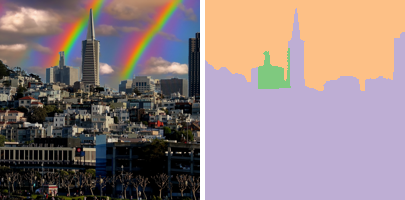}
        \caption{Three-source}
    \end{subfigure}
    \hfill 
    \begin{subfigure}[b]{0.32\linewidth}
        \centering
        \includegraphics[width=\linewidth, height=1.46cm]{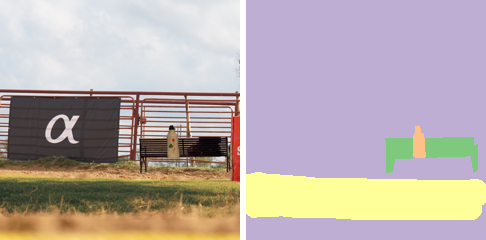} 
        \caption{Four-source}
    \end{subfigure}
    \vspace{0cm}
    
    \caption{SafireMS-Expert examples. Each color represents a single source region.}
    \label{saf.fig.ms-exp}%
\end{figure}

While SafireMS-Auto is an automatically generated dataset, SafireMS-Expert is meticulously crafted through manual manipulation. The original images are sourced from DPReview (www.dpreview.com) and COCO 2017~\cite{lin_microsoft_2015}. The forgery methods utilized in SafireMS-Expert include splicing, removal by AI-based inpainting~\cite{yu2023inpaint}, reconstructing some objects using generative models, and adding objects by generative models using text prompts~\cite{zhang2023adding}. Experts are assigned specific combinations of manipulation types to guide their image generation. They begin by selecting images that fit the specified manipulation types, then proceed to designate areas for manipulation, assign coordinates to paste, apply post-processing, choose parameters for generative models, and adjust text prompts. This process enables them to create realistic multi-source images. A total of 238 image-mask pairs are created, with details provided in Tab. \ref{saf.tab.SafireMS-Expert} and examples in Fig. \ref{saf.fig.ms-exp}.

\newcolumntype{M}{>{\centering\arraybackslash}m{1.7cm}}
\begin{table}[t]
\centering
{
\fontsize{9pt}{10pt}\selectfont
\begin{tabular}{MMMM}
\toprule
Dataset                   & Number of Sources & Images (Types)      & Total Images         \\ \midrule
\multirow{3}{*}{SafireMS-Expert} & 2                 & 32 (4) & \multirow{3}{*}{238} \\
                          & 3                 & 118 (13) &                      \\
                          & 4                 & 88 (22) &       \\
\bottomrule
\end{tabular}}
\caption{The composition of SafireMS-Expert. It can be categorized by the number of sources used for an image, and each category contains various combinations of forgery methods.}
\label{saf.tab.SafireMS-Expert}
\end{table}

\section{Visualizing Prediction Maps}
\label{saf.sec.vis_pred_maps}

\begin{figure*}[]
\centering{\includegraphics[width=0.4\linewidth]{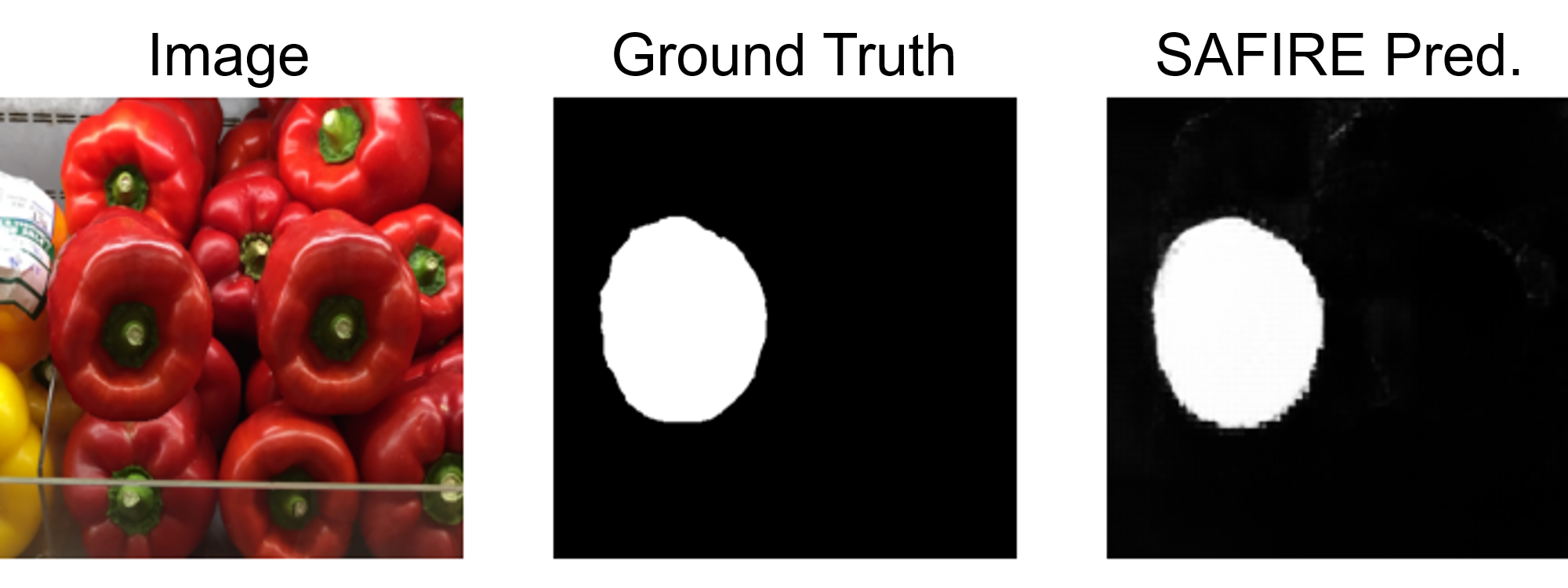}}
\centering{\includegraphics[width=1.0\linewidth]{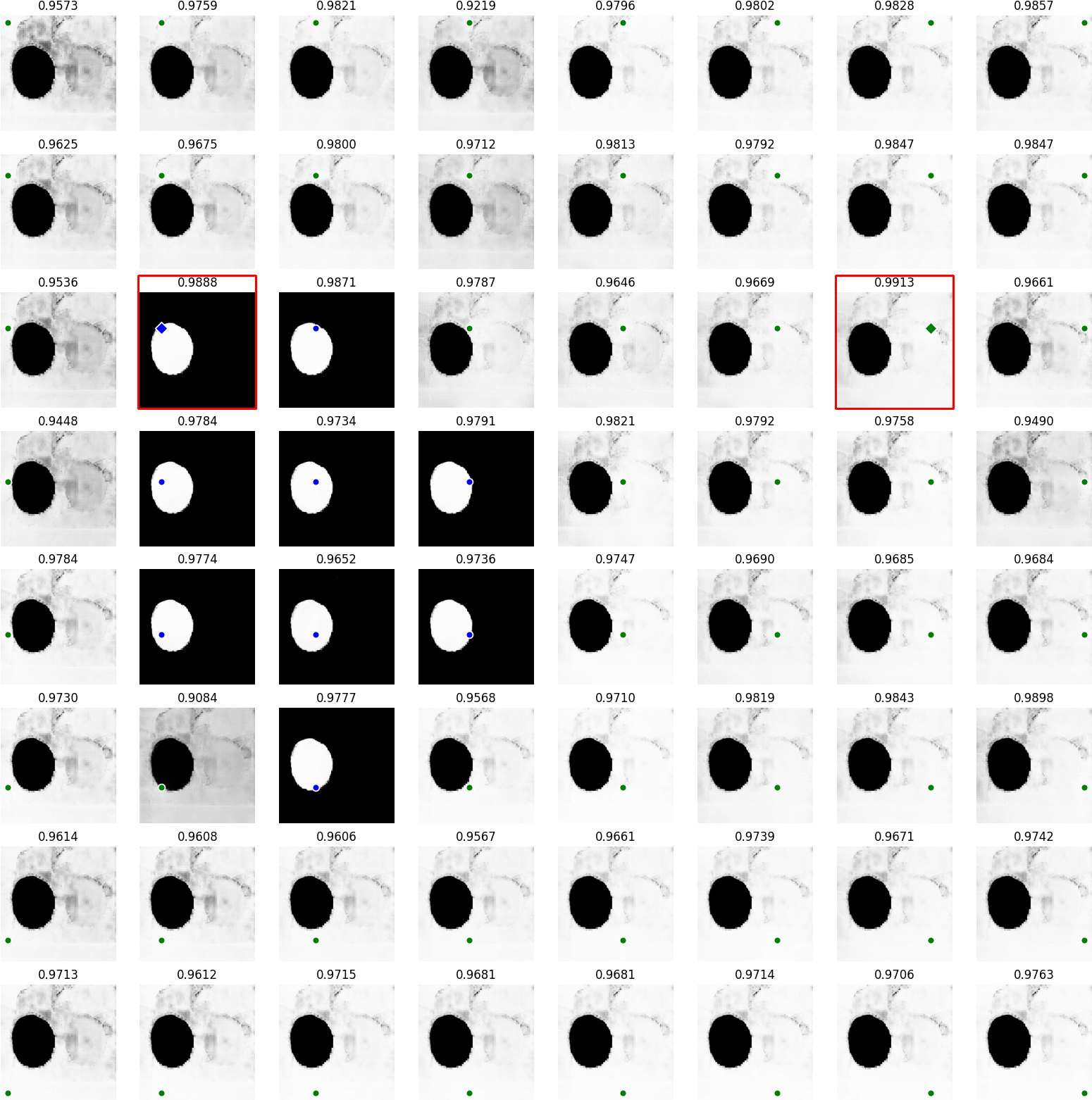}}
\caption{Prediction maps. Each image illustrates the prediction map of the SAFIRE model when an input image and a green or blue dot as a prompt are supplied. The color of the dots represents the result of clustering and the numbers at the top of the images represent the confidence scores. 
The most confident map of each cluster is highlighted with a red border.
The final prediction obtained from these two confident maps is located in the top right corner.
}
\label{saf.fig.pred_maps}
\end{figure*}

To better understand the SAFIRE inference process, this section examines the prediction maps before clustering.
Figure \ref{saf.fig.pred_maps} shows the prediction maps for each point when point prompts are given in the $8 \times 8$ grid during the inference process.
The value at the top of each image represents the confidence score of it.

The results indicate that the areas containing each point prompt are correctly segmented as 1 (white), while the other areas are marked as 0 (black).
Additionally, the color of the points (green or blue) represents the clustering results. As expected, points located on the forged areas form one cluster, while those outside form another cluster.
The map with the highest confidence score from each cluster contributes to forming the final map shown in the top right corner of the figure. Since this final map resembles the ground truth, it demonstrates that SAFIRE effectively performs IFL.

\section{Additional Experiments}

When manipulated images are created and spread as fake news, they invariably undergo subtle global post-processing, such as minor compression, often unbeknownst to viewers~\cite{wu2022robust,rao2021self}. Demonstrating robust performance under such harsh conditions is a crucial requirement for an outstanding IFL model.

In our robustness test, we evaluate the IFL performance by applying four commonly used global post-processing techniques to images, following~\cite{guillaro2023trufor}. These include Gaussian blur, Gaussian noise, JPEG compression, and gamma correction. 
We use 100 images randomly chosen from the test set in the robustness test.
The test is performed on the top five models that recorded high scores in the main experiments.

\begin{figure}[h]
\centering{\includegraphics[width=1.0\linewidth]{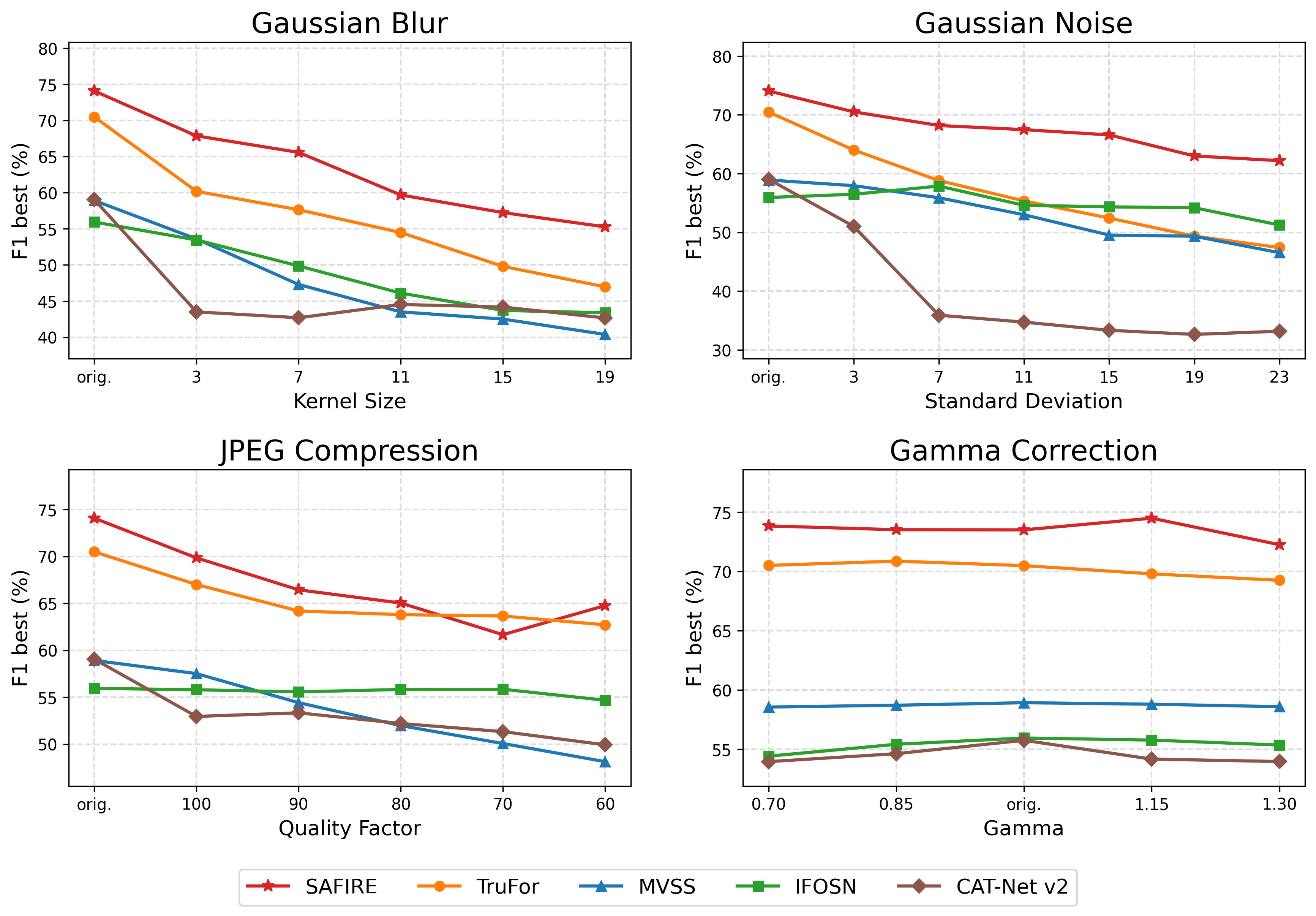}}
\caption{Robustness test. 
The performance changes of each method are measured using images subjected to four types of global post-processing.
SAFIRE demonstrates stable performance even with increased intensity of post-processing, indicating its robustness in harsh environments.
}
\label{saf.fig.robust}
\end{figure}

Figure \ref{saf.fig.robust} demonstrates that SAFIRE outperforms all other methods under all global post-processing with various parameters.
This suggests that SAFIRE is expected to show effective performance on images circulated through social media, potentially helping to mitigate the spread of fake news.

\begin{figure}[H]
\centering{\includegraphics[width=1.0\linewidth]{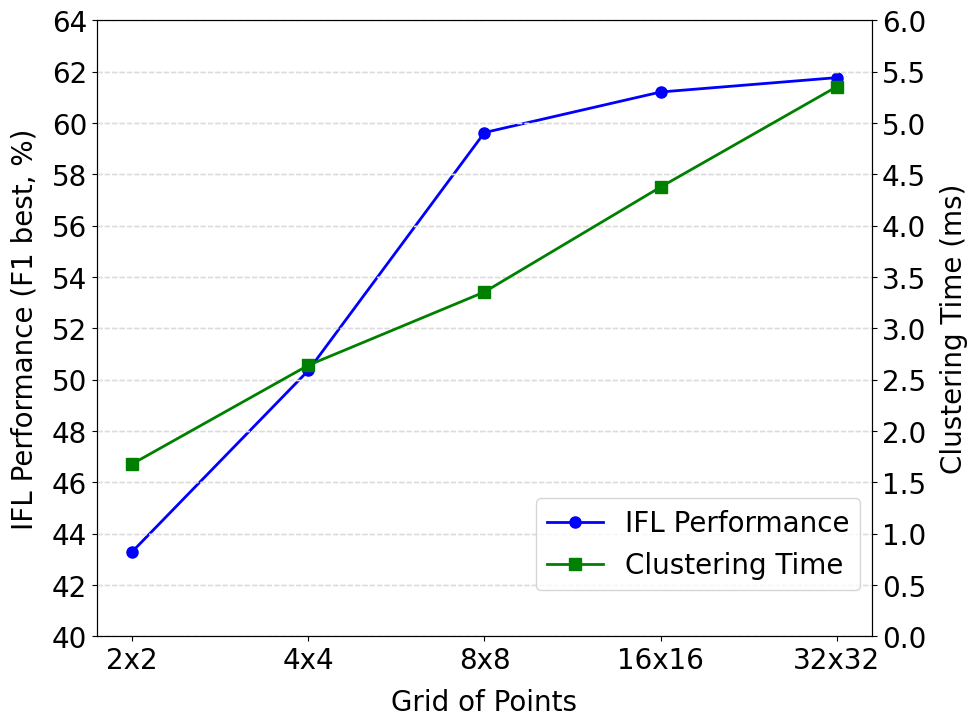}}
\caption{Effect of the number of points. 
As the points are input more densely, the IFL performance gradually improves until it saturates. At the same time, the clustering time also increases.
}
\label{saf.fig.abl_num_points}
\end{figure}

The effect of the number of points is shown in Fig.~\ref{saf.fig.abl_num_points}. 
The performance increases with denser point grids but so does the clustering time. 
At $16 \times 16$, the clustering time is sufficiently fast (0.0044 sec per image) with no significant increase in performance after that.

\begin{figure}[H]
\centering{\includegraphics[width=1.0\linewidth]{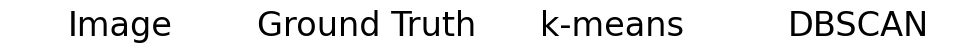}}
\centering{\includegraphics[width=1.0\linewidth]{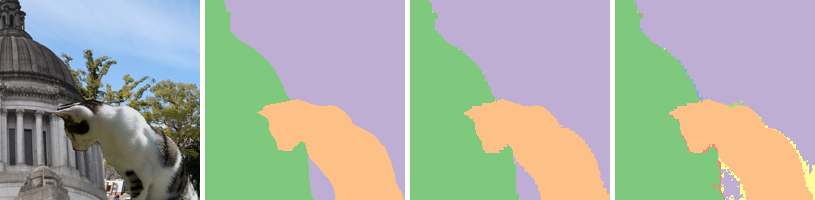}}
\centering{\includegraphics[width=1.0\linewidth]{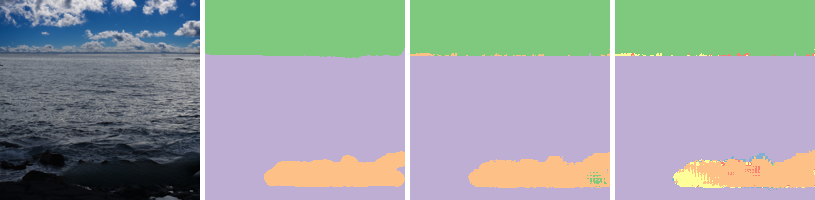}}
\centering{\includegraphics[width=1.0\linewidth]{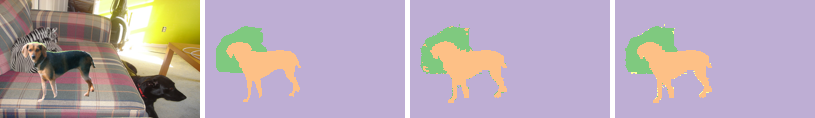}}
\centering{\includegraphics[width=1.0\linewidth]{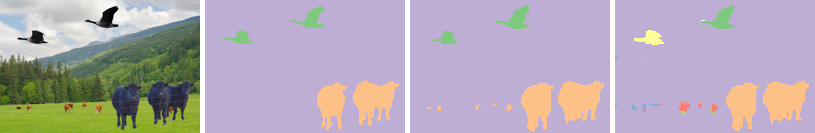}}
\centering{\includegraphics[width=1.0\linewidth]{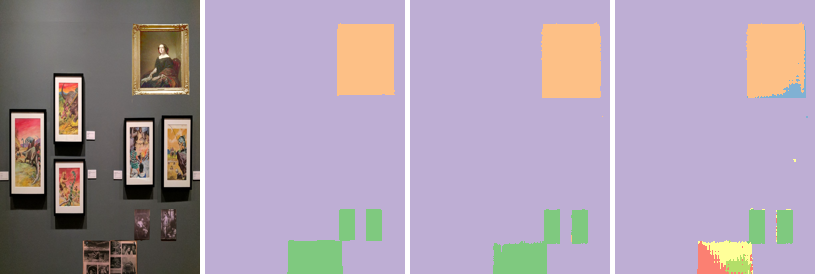}}
\centering{\includegraphics[width=1.0\linewidth]{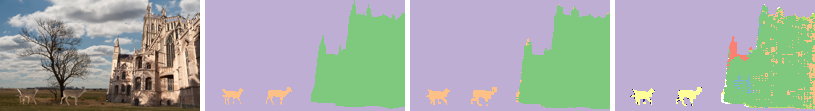}}
\centering{\includegraphics[width=1.0\linewidth]{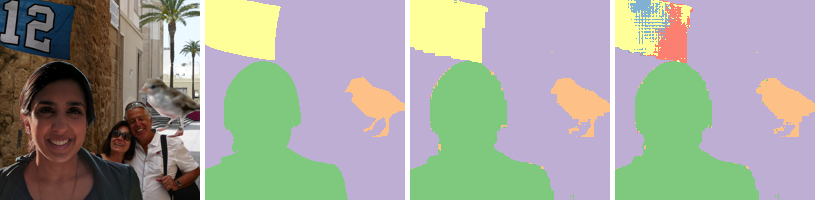}}
\centering{\includegraphics[width=1.0\linewidth]{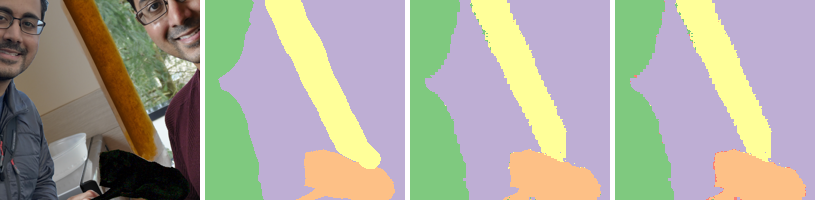}}
\centering{\includegraphics[width=1.0\linewidth]{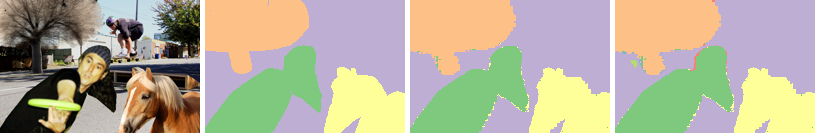}}
\centering{\includegraphics[width=1.0\linewidth]{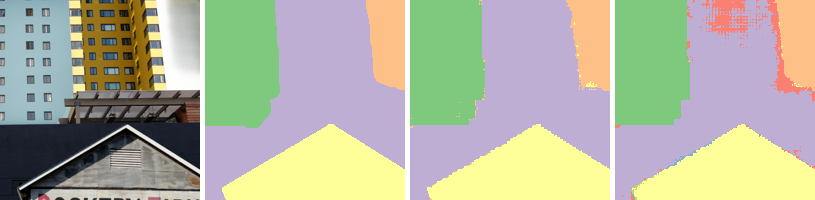}}
\centering{\includegraphics[width=1.0\linewidth]{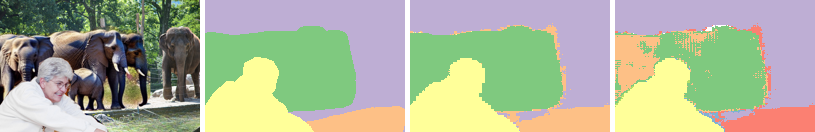}}
\centering{\includegraphics[width=1.0\linewidth]{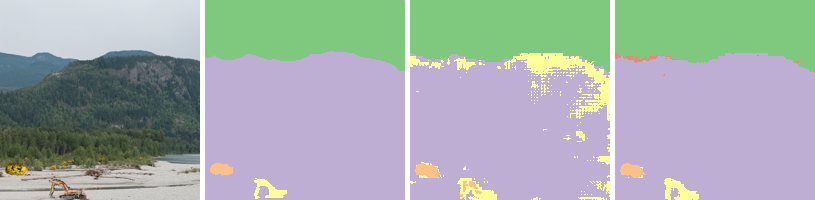}}
\caption{Visualization of multi-source partitioning. Each color represents a single source region.}
\label{saf.fig.ms_more}
\end{figure}

Additional visualizations of SAFIRE are presented in Fig.~\ref{saf.fig.ms_more}. These include the results of k-means clustering when the number of sources is known in advance and the results of DBSCAN, which predicts the number of sources automatically.

\section{Experimental Environment}

The experiments were conducted in an Ubuntu 20.04.6 LTS, CUDA 11.8, and PyTorch 2.3.0 environment. For training, six NVIDIA RTX 4090 (24GB) GPUs and Intel Xeon Gold 6348 CPU (2.60GHz) were used. For testing, SAFIRE and the comparison models were run on a single GPU in the same environment, except for some comparison methods that required more memory, where an NVIDIA A100 (80GB) was used. All results of the proposed method and comparison techniques were measured once identically. Other libraries and details required for reproducing the results are included in the official repository.

\section{Metrics for Multi-source Partitioning}
This section elaborates on the metrics for multi-source partitioning.
We use the mean Intersection over Union (mIoU) and the Adjusted Rand Index (ARI) to evaluate the multi-source partitioning performance. 

In fact, mIoU is a commonly used metric in semantic segmentation, measuring how closely the model's predictions align with the ground truth map.
The authors of~\cite{kwon2021cat} applied the concept of permuted metrics from IFL~\cite{huh_fighting_2018} to mIoU in a two-class scenario, utilizing permuted mIoU (p\_mIoU).
We generalize p\_mIoU to $N$-source partitioning with arbitrary $N$.

To recap, in traditional IFL, the permuted metric $p\_met(\cdot,\cdot)$ of a conventional metric $met(\cdot,\cdot)$ is defined as follows~\cite{huh_fighting_2018}:
\begin{align}
    p\_met(Y,X^*) = \text{max}(met(Y,X^*), met(Y,1-X^*)),
\end{align}
where $Y$ is the ground truth and $X^*$ is the prediction map defined in Eq. (10) of the main text.

In multi-source partitioning with $N$ sources, we generalize $p\_met(\cdot,\cdot)$ as follows:
\begin{align}
    p\_met(Y,X^*) = \mathop{\text{max}}_{\tilde{X}^* \in \textnormal{Perm}(X^*)} met(Y,\tilde{X}^*),
\end{align}
where $\textnormal{Perm}(\cdot)$ returns the set of all permutations in label dimension.

When $N$ is given in advance, such computation is straightforward; however, when the algorithm must predict $N$, the number of sources $N_{pred}$ inferred by the algorithm may differ from the actual number of sources $N$.
In this case, considering all permutations of the prediction map might result in excessively high computational demand.
Therefore, when $N_{pred} > N$, only the $N$ largest source regions are treated as valid predictions, while others are treated as wrong predictions. In other words, only $N$ largest source predictions are permuted to calculate the score and other predicted labels are all marked as wrong.

ARI is a measure used to evaluate the similarity between two data clusterings, adjusted to account for chance. 
It provides a value between -1 and 1, where 1 indicates perfect agreement between two clusterings, 0 indicates random clustering, and negative values indicate independent or dissimilar clusterings.
Since ARI inherently accounts for permutations automatically, it is well-suited for evaluating the performance of label-agnostic segmentation like in multi-source partitioning.

\end{appendix}

\bibliography{aaai25}

\section{Reproducibility Checklist}

This paper
\begin{itemize}
\item Includes a conceptual outline and/or pseudocode description of AI methods introduced (yes)
\item  Clearly delineates statements that are opinions, hypothesis, and speculation from objective facts and results (yes)
\item  Provides well marked pedagogical references for less-familiare readers to gain background necessary to replicate the paper (yes)

\end{itemize}

\noindent
Does this paper make theoretical contributions? (no)

\noindent
Does this paper rely on one or more datasets? (yes)

\noindent
If yes, please complete the list below.
\begin{itemize}
\item A motivation is given for why the experiments are conducted on the selected datasets (yes)
\item All novel datasets introduced in this paper are included in a data appendix. (partial)
\item All novel datasets introduced in this paper will be made publicly available upon publication of the paper with a license that allows free usage for research purposes. (yes)
\item All datasets drawn from the existing literature (potentially including authors’ own previously published work) are accompanied by appropriate citations. (yes)
\item All datasets drawn from the existing literature (potentially including authors’ own previously published work) are publicly available. (yes)
\item All datasets that are not publicly available are described in detail, with explanation why publicly available alternatives are not scientifically satisficing. (NA)
\end{itemize}

\noindent
Does this paper include computational experiments? (yes)

\noindent
If yes, please complete the list below.
\begin{itemize}
\item Any code required for pre-processing data is included in the appendix. (no).
\item All source code required for conducting and analyzing the experiments is included in a code appendix. (no)
\item All source code required for conducting and analyzing the experiments will be made publicly available upon publication of the paper with a license that allows free usage for research purposes. (yes)
\item All source code implementing new methods have comments detailing the implementation, with references to the paper where each step comes from (yes)
\item If an algorithm depends on randomness, then the method used for setting seeds is described in a way sufficient to allow replication of results. (partial)
\item This paper specifies the computing infrastructure used for running experiments (hardware and software), including GPU/CPU models; amount of memory; operating system; names and versions of relevant software libraries and frameworks. (yes)
\item This paper formally describes evaluation metrics used and explains the motivation for choosing these metrics. (yes)
\item This paper states the number of algorithm runs used to compute each reported result. (yes)
\item Analysis of experiments goes beyond single-dimensional summaries of performance (e.g., average; median) to include measures of variation, confidence, or other distributional information. (no)
\item The significance of any improvement or decrease in performance is judged using appropriate statistical tests (e.g., Wilcoxon signed-rank). (no)
\item This paper lists all final (hyper-)parameters used for each model/algorithm in the paper’s experiments. (yes)
\item This paper states the number and range of values tried per (hyper-) parameter during development of the paper, along with the criterion used for selecting the final parameter setting. (partial)
\end{itemize}

\end{document}